%% file: main.tex
\documentclass[runningheads]{llncs}

\usepackage{eccv}

\usepackage{eccvabbrv}

\usepackage{graphicx}
\usepackage{booktabs}

\usepackage[accsupp]{axessibility}  % Improves PDF readability for those with disabilities.

\usepackage{hyperref}

\usepackage{orcidlink}

\input{sections/preamble}
\begin{document}

\title{
CloudFixer: \\
Test-Time Adaptation for 3D Point Clouds \\ via Diffusion-Guided Geometric Transformation
}

\titlerunning{CloudFixer}

\author{
Hajin Shim\inst{1,2}$^*$\orcidlink{0000-0002-0116-3237} \quad
Changhun Kim\inst{1,3}$^*$\orcidlink{0009-0003-4930-6908} \quad
Eunho Yang\inst{1,3}\orcidlink{0000-0003-2188-0169}
}

\authorrunning{Shim et al.}

\institute{
Korea Advanced Institute of Science and Technology (KAIST), South Korea \and
Samsung Advanced Institute of Technology (SAIT), South Korea \and
AITRICS, South Korea \\
\email{\{shimazing, changhun.kim, eunhoy\}@kaist.ac.kr}
}

\maketitle
\def\thefootnote{*}\footnotetext{Equal contribution.}

\input{sections/abstract}
\input{sections/introduction}

\input{sections/related_work}
\input{sections/preliminary}
\input{sections/method}
\input{sections/experiments}
\input{sections/conclusion}
\input{sections/acknowledgement}

% \clearpage % TODO REVIEW/FINAL: This \clearpage needs to be removed from both review and camera-ready versions.

\bibliographystyle{splncs04}
\bibliography{main}

\input{sections/appendix}

\end{document}

%% file: sections/preamble.tex
\usepackage[dvipsnames]{xcolor}
\usepackage[utf8]{inputenc} % allow utf-8 input
\usepackage[T1]{fontenc}    % use 8-bit T1 fonts
\usepackage{url}            % simple URL typesetting
\usepackage{booktabs}       % professional-quality tables
\usepackage{makecell}
\usepackage{amsfonts}       % blackboard math symbols
\usepackage{nicefrac}       % compact symbols for 1/2, etc.
\usepackage{microtype}      % microtypography
\usepackage{amsmath}
\usepackage{graphicx}
\usepackage{amssymb}
\usepackage{caption}
\usepackage{subcaption}
\usepackage{bbm}
\usepackage{eqparbox}
\usepackage{arydshln}
\usepackage{multirow}
\usepackage{pifont}
\usepackage{float}
\usepackage{wrapfig}
\usepackage{makecell}
\usepackage{verbatim}
\usepackage{tabularray}
\usepackage{kotex}
\usepackage{algorithm}
\usepackage{graphicx} % for resizebox
\usepackage{adjustbox}
\usepackage[noend]{algpseudocode}
\usepackage{array}
\usepackage{color}
\usepackage{colortbl}
\usepackage{xr-hyper}

\newcolumntype{P}[1]{>{\centering\arraybackslash}p{#1}}

\newcommand*\colourcheck[1]{
  \expandafter\newcommand\csname #1check\endcsname{\textcolor{#1}{\ding{51}}}%
}
\newcommand*\colourx[1]{
  \expandafter\newcommand\csname #1x\endcsname{\textcolor{#1}{\ding{55}}}%
}

\newcommand{\LineComment}[1]{\textcolor{lightgray}
{%\State 
$\triangleright \text{#1}$}}

\definecolor{lightgray}{rgb}{0.7, 0.7, 0.7}
\definecolor{custompink}{rgb}{0.99, 0.03, 0.51}
\colourx{lightgray}
\colourx{custompink}

%% file: sections/abstract.tex
\begin{abstract}
3D point clouds captured from real-world sensors frequently encompass noisy points due to various obstacles, such as occlusion, limited resolution, and variations in scale. These challenges hinder the deployment of pre-trained point cloud recognition models trained on clean point clouds, leading to significant performance degradation. While test-time adaptation (TTA) strategies have shown promising results on this issue in the 2D domain, their application to 3D point clouds remains under-explored. Among TTA methods, an input adaptation approach, which directly converts test instances to the source domain using a pre-trained diffusion model, has been proposed in the 2D domain. Despite its robust TTA performance in practical situations, naively adopting this into the 3D domain may be suboptimal due to the neglect of inherent properties of point clouds, and its prohibitive computational cost. Motivated by these limitations, we propose CloudFixer, a test-time input adaptation method tailored for 3D point clouds, employing a pre-trained diffusion model. Specifically, CloudFixer optimizes geometric transformation parameters with carefully designed objectives that leverage the geometric properties of point clouds. We also substantially improve computational efficiency by avoiding backpropagation through the diffusion model and a prohibitive generation process. Furthermore, we propose an online model adaptation strategy by aligning the original model prediction with that of the adapted input. Extensive experiments showcase the superiority of CloudFixer over various TTA baselines, excelling in handling common corruptions and natural distribution shifts across diverse real-world scenarios. Our code is available at \url{https://github.com/shimazing/CloudFixer}.

\end{abstract}

%% file: sections/introduction.tex
\section{Introduction}
\label{sec:intro}
Recent advancements in 3D vision have established point clouds as expressive representations of the natural world~\cite{guo2020deep}, enabling lots of applications ranging from autonomous driving~\cite{chen20203d} to augmented reality~\cite{alexiou2017towards}. These advancements have been achieved by the development of various deep neural networks tailored for point cloud recognition~\cite{dgcnn,pointnet,pointnet++,pointMAE,point2vec}. These models are primarily trained on well-curated clean benchmark datasets; however, in real-world scenarios, point clouds collected from physical devices, such as LiDAR sensors, often encompass noisy points due to multiple factors, including occlusion, fluctuation in scale and density, limited resolution, and vibration or motion of the capturing devices. % This \emph{distribution mismatch between} source and target domain poses challenges in deploying point cloud recognition models trained on carefully curated clean point clouds to real-world applications as any misinterpretation of this data could potentially lead to catastrophic consequences.
The distribution mismatch between the source and target domain presents challenges in deploying point cloud recognition models %trained on carefully curated clean point clouds 
to real-world applications, as any misinterpretation of this data could potentially lead to catastrophic consequences.

% \B{Addressing the distribution shift problem is of paramount importance in the 3D domain, especially in applications such as autonomous driving, where human lives are at stake.}
To address distribution shift problems, a novel approach called test-time
\begin{wrapfigure}{r}[0in]{0.3\textwidth}
    % \vspace{-.3in}
    \includegraphics[width=0.3\textwidth]{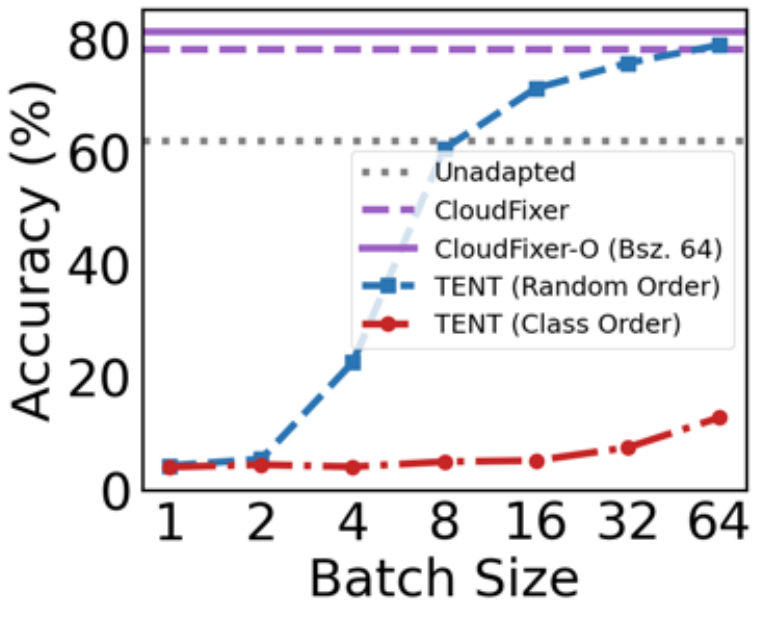}
    % \vspace{-.3in}
    \caption{Accuracy of TENT~\cite{wang2020tent} and CloudFixer across different batch sizes and label distributions (Random Order vs. Class Order).}
    \label{fig:motivation}
    % \vspace{-.3in}
\end{wrapfigure}
adaptation (TTA)~\cite{pl, shot, wang2020tent, sun2020test, liu2021ttt++, gao2022source, memo, mirza2022mate, lame, note, dua, sgem, niu2023towards} has recently emerged to adapt a pre-trained model to an arbitrary target domain during the inference phase in an unsupervised manner. % TTA aims to adapt a pre-trained model from the source domain to an arbitrary target domain during the inference phase in an unsupervised manner without access to the source data.
Despite the abundance of TTA research in 2D vision and the necessity for developing effective TTA approaches for the 3D domain, there has been only limited investigation into TTA methods for 3D point clouds~\cite{mirza2022mate}. Furthermore, we unveil upon comprehensive investigation that naively applying conventional 2D test-time model adaptation approaches does not yield enough performance gain; this issue is particularly pronounced in realistic scenarios with constraints like limited batch sizes, temporally correlated test streams, and label distribution shifts, as depicted in~\cref{fig:motivation}. It is primarily attributed to the heavy reliance on \emph{unstable model predictions} while underscoring the necessity for novel TTA approaches over conventional model adaptation methods.

In response, a method called DDA~\cite{gao2022source} was proposed to address these issues in the 2D vision domain. Instead of adapting model parameters to fit the target domain, DDA adapts the input to match the source domain using the domain translation capabilities of a pre-trained diffusion model~\cite{ilvr, sdedit}. This involves forward process of adding Gaussian noise to an image, followed by iterative reverse process that aligns the image to the source domain while preserving overall shape integrity. Its per-sample basis manner operation robustness in real-world scenarios, such as limited batch sizes, temporally correlated test streams, and label distribution shifts. However, transitioning DDA to the 3D domain may be suboptimal due to its oversight of the geometric properties of point clouds and its prohibitive computational cost for real-time applications.

Motivated by these limitations, we introduce \emph{CloudFixer}, a test-time input adaptation method tailored for 3D point clouds, leveraging a pre-trained diffusion model.
Specifically, adaptation is achieved through optimization steps learning rotation matrix parameters and point displacements guided by the source diffusion model. We also suggest a per-point regularization weight for flexibility with noisy isolated points while preserving core information. To address computational costs, our update procedure ensures that the diffusion model performs only a forward pass without backpropagation during adaptation. While our primary focus is achieving robust test-time adaptation in realistic scenarios via instance-wise input adaptation, we further propose an online model adaptation technique, minimizing a consistency loss to align class predictions of adapted and original inputs.
Our experiments demonstrate that CloudFixer achieves state-of-the-art performance in various distribution shift scenarios, encompassing common and realistic corruptions in the ModelNet40-C~\cite{modelnet40c} benchmark and natural distribution shifts in PointDA-10~\cite{qin2019pointdan}. To summarize, our contributions are threefold:
\begin{itemize}
    \item We propose CloudFixer, which is the first test-time input adaptation strategy tailored for 3D point clouds, proposing domain-specific parameterization and objective, harnessing pre-trained diffusion models.

    \item CloudFixer is well-suited for real-time 3D applications as it requires neither backpropagation through the diffusion model nor an excessive generation process, enabling it to adapt a single instance in less than 1 second.

    \item Through extensive experiments, we demonstrate that our method achieves state-of-the-art performance across diverse distribution shift scenarios, encompassing common corruptions and natural distribution shifts. % across various classifier architectures.
\end{itemize}

%% file: sections/related_work.tex
\section{Related Work}
\subsubsection{Domain Adaptation and Generalization on Point Clouds}
% general concept -> unsupervised domain adaptation -> domain generalization -> limitation
Given the frequent exposure of point clouds to distribution shifts in real-world scenarios, various domain adaptation and generalization strategies have been proposed. % to improve the generalization performance of point cloud recognition models trained on the source domain.
A significant branch involves unsupervised domain adaptation (UDA)~\cite{qin2019pointdan, achituve2021self, alliegro2021joint, shen2022domain, fan2022self, cardace2023self, zou2021geometry}. These methods aim to achieve compatible performance on the target domain by leveraging labeled source data and unlabeled target data. They typically employ unsupervised objectives, such as pseudo-labeling %based on the main task of the target domain
~\cite{fan2022self}, and self-supervised tasks that predict the geometric properties %of given point clouds
~\cite{qin2019pointdan, achituve2021self, shen2022domain, zou2021geometry, cardace2023self, alliegro2021joint}.
Another big category is domain generalization (DG)~\cite{lehner20223d, wei2022learning, xiao2022learning, wei2022learning}. %, which aims to enhance models' generalization capability across diverse domains. 
These primarily involve adversarial learning or explicit feature alignment %to ensure alignment of features 
across multiple domains~\cite{wei2022learning, xiao2022learning, huang2022manifold} and utilize data augmentation methods~\cite{lehner20223d}.
Despite showcasing decent performance gains, these methods have inability to operate in cases where source data is inaccessible due to privacy or storage concerns. %, and a lack of adaptability to tailor the model to the current test data.

\subsubsection{Test-Time Adaptation}
% general concept -> three types: TTT, fully tta, bn calibration -> limitation
Test-time adaptation (TTA)~\cite{pl, shot, wang2020tent, sun2020test, liu2021ttt++, gao2022source, memo, mirza2022mate, lame, note, dua, sgem, niu2023towards} strategies have emerged to address limitations in UDA/DG. % and domain adaptation strategies under distribution shift scenarios. 
These methods aim to adapt pre-trained models from a source %domain 
to an arbitrary target domain on the fly, utilizing only unlabeled target data without access to source data. TTAs are typically categorized as follows. %ully test-time adaptation, test-time training, and batch-norm statistics calibration. 
Firstly, fully TTA~\cite{pl, shot, wang2020tent, memo, niu2023towards, sgem} involves training the model unsupervisedly, leveraging self-training on the unlabeled target dataset. %without modifying the training process. 
Secondly, test-time training~\cite{sun2020test, liu2021ttt++, mirza2022mate} entails training the model with both the main objective and an additional self-supervised task and utilizing this self-supervision for adaptation during inference. Next, batch-norm statistics calibration methods~\cite{dua, bn_stats, lim2023ttn, note} update statistics of batch normalization layers using test instances to estimate unbiased normalization statistics of target domain.
In contrast, our approach diverges from these methods by projecting input instances into the training data regime as \cite{gao2022source} using a pre-trained diffusion model on the source domain tailored for point clouds.

\subsubsection{Diffusion Models}
% general concept -> diffusion model for point clouds -> diffusion-guided domain translation -> ours
Diffusion models~\cite{song2020score, ddpm, ddim}, which approximate the \emph{reverse} of the diffusion process to learn the training manifold, have gathered significant attention as prominent generative models. They have been widely used in various domains, including 2D/3D vision~\cite{ramesh2021zero, ramesh2022hierarchical, saharia2022photorealistic, zhou20213d, zeng2022lion, liu2023meshdiffusion, jun2023shape}, videos~\cite{ho2022video, ho2022imagen, kim2022diffusion}, and languages~\cite{li2022diffusion, gong2022diffuseq}.
For 3D point clouds, %several works about 
several diffusion models have been proposed, especially for generation~\cite{zhou20213d, e3_diffusion, Luo_2021_CVPR}, completion~\cite{lyu2021conditional, ma2022unsupervised}, and manipulation~\cite{zeng2022lion}. %These models also facilitate the conditional generation of point clouds using image, text or shape latent representations~\cite{luo2021diffusion, jun2023shape}.
Diffusion models exhibit remarkable abilities in translating the domain of given inputs to the source domain. For example, ILVR~\cite{ilvr} accomplishes domain translation by iteratively generating an image while preserving the low-pass filtering results, and SDEdit~\cite{sdedit, egsde} dilutes the reference image of a different domain through a forward process, projecting it into the latent space where the two domains intersect. %the intersection of the two domains occurs.
%Our approach relies on diffusion-guided domain translation, departing from generation-based translation techniques commonly used. 
Instead of these, % generation-based methods,
we adopt an optimization-based method~\cite{poole2022dreamfusion}, while acknowledging the unique characteristics of point clouds. %This involves learning parameterized transformations of point clouds to better adapt to the target domain.

%% file: sections/preliminary.tex
\section{Preliminary}
\label{sec:preliminary}
% \B{
% In this section, we delineate our target problem in \Cref{subsec:problem} and provide a brief explanation of diffusion models in \Cref{subsec:diffusion_model}, which play a crucial role in guiding the source domain in our method.
% }

\subsection{Problem Setup}
\label{subsec:problem}
Let $q(x)$ be a source distribution of point clouds $x \in \mathbb{R}^{N\times3}$ consisting of $N$ points for training dataset $\mathcal{D}_s = {\{(x_i^s, c_i^s)\}}_i$ in pairs of point clouds and class labels, and $f_\psi(x) \in \mathbb{R}^C$ be a $C$-class classification model trained on $\mathcal{D}_s$. Given unlabeled target domain $\mathcal{D}_t = {\{ x_i^t \}}_i$ under distribution shifts, \ie, $ x_i^t \not\sim q(x)$, our test-time input adaptation aims to achieve robust prediction performance under distribution shifts. Existing TTA methods usually adapt the model $f_\psi$ by optimizing $\psi$ using self-training, 
based on the prediction on test instances or replacing the statistics %of the training data
in normalization layers with those of test examples~\cite{shot, wang2020tent, memo, niu2023towards}. Instead, we leverage the domain translation ability of a pre-trained diffusion model to directly transform the input into the source domain.

\subsection{Diffusion Models}
\label{subsec:diffusion_model}
Diffusion models~\cite{song2020score, ddpm, ddim} are generative models that estimate the reverse process of the diffusion process to generate the data $x_0 \sim q(x_0)$ by gradually denoising random noise $x_T \sim \mathcal{N}(x_T; 0, I)$. For each timestep $t=0, \cdots, T$, the marginal distribution of $x_t$ given $x_0$ is defined as $q(x_t|x_0) = \mathcal{N}(x_t; \alpha_t x_0, \sigma_t^2 I)$, where $\alpha_t$ strictly decreases from 1 to 0 as $t$ increases, and % $\alpha_t^2 + \sigma_t^2 = 1$.
$\sigma_t = \sqrt{1 - \alpha_t^2}$.
The models are trained for $p_\theta(x_{t-1}|x_t)$ to approximate 
%The objective is %These models are trained 
%to approximate 
$q(x_{t-1} | x_t, x_0)$ by minimizing %by $p_\theta(x_{t-1}|x_t)$: % as follows: %using the following objective:
\begin{equation*}
    \mathcal{L}_{\texttt{diff}}(\theta) = \mathbb{E}_{x_0\sim q(x_0),\epsilon\sim\mathcal{N}({0},{I}),t} \Big[w(t)\|\epsilon_\theta(x_t,t)-\epsilon\|_2^2\Big],
\end{equation*}
where $w(t)$ is a weight function of timestep $t$,  $x_t = \alpha_t x_0 + \sigma_t \epsilon$ is the forwarded input, and the parameterized model $\epsilon_\theta(\cdot, t)$ is to predict truly injected Gaussian noise $\epsilon$ in the forward process. With the estimated noise, %into the equation of the forward process and conducting rearrangement, 
we can estimate the original data point $\hat{x}_0$ from $x_t$ as follows:
\begin{equation*}
    \hat{x}_0 = \frac{x_t - \sigma_t \epsilon_\theta(x_t, t)}{\alpha_t}. 
\end{equation*}
Here, $\epsilon_\theta(x_t, t)$, which approximates $-\sigma_t\nabla_{x_t}q(x_t)$ allows us to obtain the direction to move from a noised input towards the data distribution. When %the initial sample 
$x_0$ is not from the source domain but from another shifted domain, the disparity between the estimation and the data point serves as guidance toward the source domain.

%% file: sections/method.tex
\section{Proposed Method: CloudFixer} %\B{: Diffusion-Guided Point Cloud Adaptation}}
\label{sec:method}
In this section, we introduce CloudFixer, which is a per-sample diffusion-guided test-time input adaptation strategy tailored for 3D point clouds. 
We outline our input adaptation and an optional online model adaptation methods in~\Cref{subsec:overview}. Following this, we cover the details of components of CloudFixer in \Cref{subsec:method_detail} and \ref{subsec:model_adapt}. The overall procedure is depicted in~\Cref{alg:cloudfixer} in \Cref{sec:alg}.

\begin{figure*}[!t]
  \centering
  \includegraphics[width=0.9\textwidth]{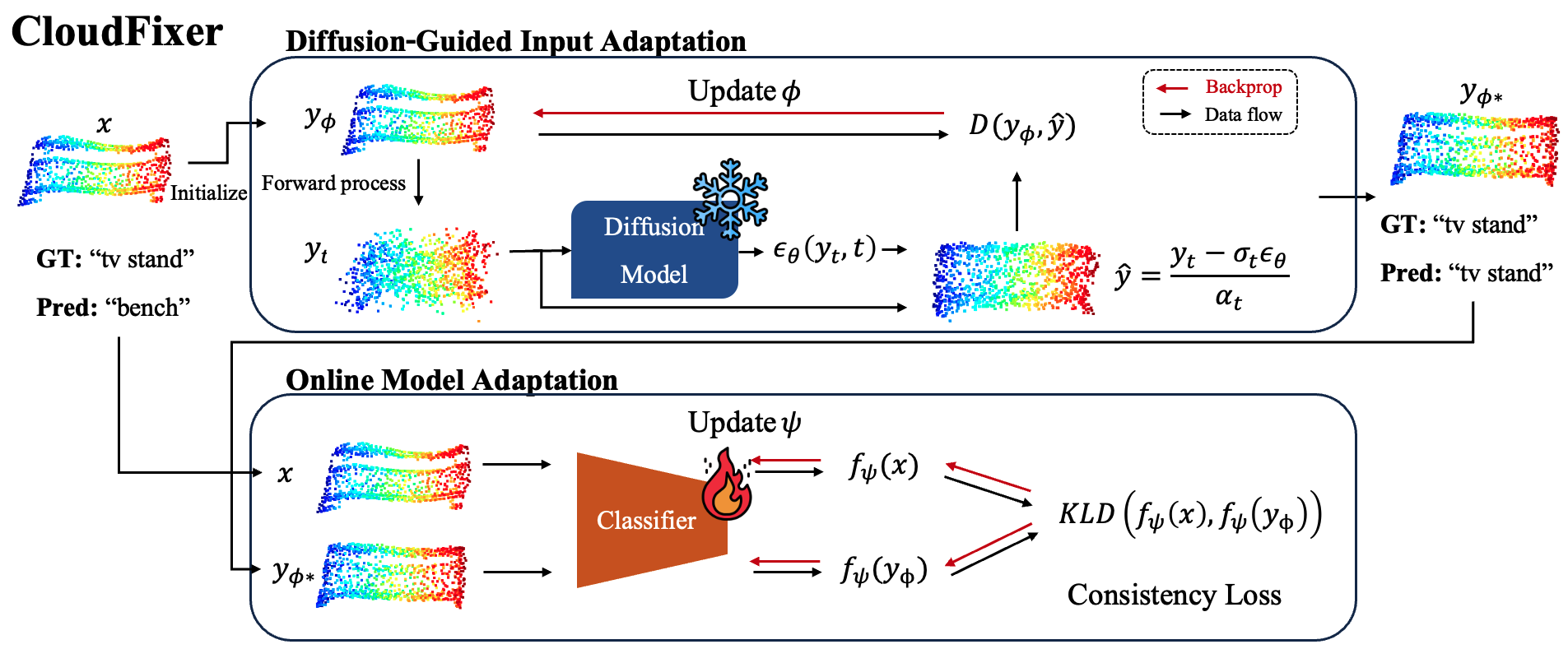}
  % \vspace{-.2in}
  \caption{CloudFixer is an optimization-based diffusion-guided input adaptation, tailored for 3D point clouds. CloudFixer iteratively optimizes geometric transformation parameters $\phi$ for $x$ to minimize the Chamfer distance between a parameterized point cloud $y_\phi$ and the estimation $\hat{y}$ from the diffusion model, aligned with the source domain. A distorted input point cloud $x$, leading to a misclassification, is transformed into $y_{\phi*}$, correcting its prediction. Additionally, online model adaptation minimizes the KL-divergence between class predictions of the original and adapted point clouds.
  }
  \label{fig:concept}
  % \vspace{-.2in}
\end{figure*}

\subsection{Overview}
\label{subsec:overview}

CloudFixer directly aligns the distribution shift of a given test instance with the source domain, leveraging the knowledge of the pre-trained diffusion model $p_\theta(x)$ on the source domain.
A straightforward domain translation using the diffusion model is a generation-based method, where data points undergo a forward process, mapping them to a noisy latent space at time $t$, followed by a denoising process \cite{sdedit, egsde, ilvr, gao2022source}. However, at larger timestep $t$, there is a risk of losing class information, while a small $t$ may not provide sufficient translation.

Building upon the limitations above, we propose novel \emph{optimization-based geometric transformation} with diffusion guidance (upper part of \cref{fig:concept}) as follows. First, we define geometric transformation $y_\phi(x)$ parameterized by $\phi$ of a test input $x$. $\phi$ is initialized for $y_\phi(x)$ to be same as $x$. We iteratively update $\phi$ with diffusion guidance.
In each iteration, we randomly sample $t$ from $U[t_{min}, t_{max}]$ and conduct a forward process from $y_\phi(x)$ to $y_t$. Subsequently, we obtain the estimation of the denoised point cloud $\hat{y}$ from the pre-trained diffusion model $\epsilon_\theta$. Since the diffusion model is trained on the source domain, $\hat{y}$ transitions from $y_\phi$ to the source domain. Therefore, we can utilize $y_\phi$ as a supervision to update $\phi$. Through this iterative optimization, we can effectively encourage stable and gradual translation of shifted test instances into the source domain.

Although our primary focus is on input adaptation, we further extend CloudFixer to include online model adaptation (CloudFixer-O) to enhance adaptation performance under conditions where enough batch size and an independent and identically distributed (i.i.d.) distributions are ensured. Specifically, we introduce a task that aligns the class distribution predictions of the original $x$ and the transformed $y_{\phi*}$ (lower part of \cref{fig:concept}). This approach facilitates domain adaptation by establishing a correlation between source and target domains, mitigating the need for uncertain information like pseudo-labels as long as the diffusion-guided transformation is valid.

\subsection{Diffusion-Guided Input Adaptation}
\label{subsec:method_detail}

\subsubsection{Parameterization of Geometric Transformation $\boldsymbol{y_\phi}$}
\label{subsec:parameterization}
We set the parameters $\phi = (R, \Delta)$ for transforming the given input $x$ as $y_\phi(x) = (x +\Delta)R^\top$ where $R \in \mathbb{R}^{3\times3}$ is a rotation matrix and $\Delta \in \mathbb{R}^{N\times3}$ is a displacement matrix of all $N$ points whose $j$-th row $\delta_j^\top$ corresponds to each point $j$. We include the rotation transformation $R$ because misalignment is a common test-time corruption of point clouds. Furthermore, the simple per-point displacement $\Delta$ is to allow flexible transformation in response to various distribution shifts.
Note here that the rotation matrix $R = [r_1; r_2; r_3]$ is further parameterized by a 6D vector $(a_1, a_2) \in \mathbb{R}^3 \times \mathbb{R}^3$ to satisfy the condition of rotation matrices, as derived by the following operation as in \cite{6drepnet}:
\begin{equation*}
    r_1 = \frac{a_1}{{\|a_1\|}_2}, \quad r_2 = \frac{u_2}{{\|u_2\|}_2}, \quad u_2 = a_2 - (r_1 \cdot a_2)r_1, \quad \text{and} \quad r_3 = r_1 \times r_2.
\end{equation*}

\subsubsection{Objective}
\label{subsec:objective}
As we mentioned in \Cref{subsec:diffusion_model}, a pre-trained diffusion model provides guidance toward the source domain of the noised input. For each iteration, after perturbing $y_\phi$ with the forward process at time $t$ as $y_t = \alpha_t y_\phi + \sigma_t \epsilon$, we estimate the denoised point cloud $\hat{y} = (y_t - \sigma_t \epsilon_\theta(y_t, t)) / \alpha_t$. The estimation moves from $y_\phi$ towards the source domain by the diffusion model. Therefore, we update the parameters to reduce the distance $D$ between $y_\phi$ and the diffusion model's estimation $\hat{y}$ as follows:
\begin{equation}
\label{eq:objective}
    \phi \leftarrow \phi - \eta \Big(\nabla_{y_\phi}D(\hat{y}, y_\phi)\frac{\partial y_\phi}{\partial\phi} + \lambda \nabla_\phi\texttt{Reg}(\phi) \Big),
\end{equation}
where $\eta$ is a learning rate. When the distance is the square of $\ell_2$ distance ${\|y - y_\phi\|}_2^2$, this update is equivalent to the Score Distillation Sampling (SDS) loss~\cite{poole2022dreamfusion} up to the scaling of each timestep $t$. However, we observe that optimizing this objective does not lead to a stable convergence towards the source domain. Instead, by leveraging the characteristic of point clouds as unordered sets,
we use the Chamfer distance to account for the permutation-invariant nature of 3D point clouds defined as follows:
\begin{equation*}
    D(x, y)  = \frac{1}{|x|}\sum_{i=1}^{|x|} \min_j 
    \Big\|x(i) - y(j) \Big\|_2^2 + \frac{1}{|y|}\sum_{j=1}^{|y|} \min_i \Big\|x(i) - y(j) \Big\|_2^2,
\end{equation*}
where $x(i)$ and $y(j)$ are 3D coordinates of point $i$ and $j$ of $x$ and $y$, respectively. It is worth noting that in CloudFixer, the diffusion model is used solely for predicting $\hat{y}$ for the supervision of $y_\phi$, and it does not involve backpropagation through the diffusion model. This stands in stark contrast to the 2D input adaptation-based TTA method DDA~\cite{gao2022source}, which requires backpropagation through the diffusion model for guidance. This key difference is crucial, especially in the 3D domain where efficiency is essential. The originality of CloudFixer compared to DDA is summarized in \Cref{sec:orig_over_dda}.

\subsubsection{Regularization}
\label{subsec:regularization}
To simply regulate excessive changes in the point cloud, we introduce a novel regularization objective by penalizing the squared norm of the displacement $\delta_j$ for each point $j$, as  $\texttt{Reg}(\Delta) = \sum_j w_j {||\delta_j||}^2_2$. Here, for each point, we calculate weights $\{w_j\}_j$ by taking the inverse of the average distance to its k-nearest neighbors, providing greater flexibility for noisy isolated points.

\subsubsection{Voting}
Due to the stochasticity of the input adaptation, we can obtain $K$ different transformations $\{y_{\phi_j}(x)\}_{j=1}^K$ of a given input $x$. We enhance classification performance by averaging the multiple predictions $\sum_j f_\psi(y_{\phi_j}) / K$, where $f_\psi(\cdot)\in [0, 1]^C $ is a model prediction of class probability.

\subsection{Online Model Adaptation}
\label{subsec:model_adapt}

We extend CloudFixer to its online version, \emph{CloudFixer-O}, by introducing a model adaptation process that incorporates the adapted inputs $\{y_{\phi_j}(x)\}_{j=1}^K$ as well as the original $x$. We perform an $M$-step update on the model except for a classification head by minimizing the KL-divergence between the class probability distribution of the test input $x$ and its adapted inputs $\{y_{\phi_j}\}$ as follows:
\begin{equation*}
\min_\psi \sum_{j=1}^K KL \left( f_\psi(x) \Big{|} f_\psi\big(y_{\phi_j}(x)\big) \right),
\end{equation*}
where $f_\psi(\cdot)\in [0, 1]^C $ is model prediction of class probability.  
Intuitively, our objective guides the feature encoder of the classifier to extract class information, disregarding shifted features, by aligning the class prediction of the test input with the source-closer adapted input.

%% file: sections/experiments.tex
\section{Experiments}
\label{sec:experiment}
This section rigorously and thoroughly demonstrates the empirical efficacy of CloudFixer. We begin by providing a detailed description of our experimental setup in \Cref{subsec:exp_setup}, followed by an elucidation of the fundamental research questions as follows: Does CloudFixer consistently outperform baselines under challenging yet realistic scenarios such as limited batch size, temporally correlated test streams, and label distribution shifts, as well as under mild conditions across various benchmarks and classifier architectures? (\Cref{subsec:main} and %\nameref{subsubsec:bsz} in 
\Cref{subsubsec:bsz}) Do the components of CloudFixer truly contribute to performance enhancement, and are they optimal choices? (%\nameref{subsubsec:ablation} in 
\Cref{subsec:ablation}) Moreover, when visualized, does CloudFixer genuinely transform point clouds into the desired source domain? Does CloudFixer demonstrate strengths in computational efficiency %(\nameref{subsubsec:efficiency} in 
(\Cref{subsubsec:efficiency}) and 
hyperparameter sensitivity %(\nameref{subsubsec:hparam_sensitivity} 
(\Cref{subsubsec:hparam_sensitivity}), which can be pivotal at test time?

\subsection{Experimental Setup}
\label{subsec:exp_setup}
\subsubsection{Datasets}
We evaluate our method on \href{https://github.com/jiachens/ModelNet40-C}{ModelNet40-C}~\cite{modelnet40c} which includes the various types common corruptions of ModelNet40~\cite{modelnet40}. It suggests 15 common and realistic corruptions that are categorized into 3 types---density, noise, and transformation. We abbreviate the corruption names as follows: OC (Occlusion), LD (LiDAR), DI (Density Inc.), DD (Density Dec.), CO (Cutout), UNI (Uniform noise), GAU (Gaussian noise), IP (Impulse), US (Upsampling), BG (Background noise), ROT (Rotation), SH (Shear), FFD (Distortion), RBF (Distortion with RBF Kernel), and IR (Distortion with Inverse RBF Kernel). We also utilize a domain adaptation benchmark, PointDA-10~\cite{qin2019pointdan}, which provides natural distribution shifts of sim-to-real and vice versa by pairing two out of the three datasets---ModelNet~\cite{modelnet}, ShapeNet~\cite{shapenet}, and ScanNet~\cite{scannet}---as the source and target domains. % It is composed of 10 classes shared across the three datasets. 
ModelNet and ShapeNet are generated from 3D CAD models, while ScanNet is created by scanning real scenes. Further details are in \Cref{sec:dataset_details}.

% \vspace{-0.1in}
\subsubsection{Model Architectures}
For classifiers, we employ Point2Vec~\cite{point2vec}, which has demonstrated notable success on ModelNet40~\cite{modelnet40}, as a classifier backbone for ModelNet40-C, by using a publicized checkpoint in the \href{https://github.com/kabouzeid/point2vec}{official repository}.
We additionally evaluate CloudFixer's performance on PointMLP~\cite{pointMLP}, PointNeXt~\cite{pointNeXt}, and PointMAE~\cite{pointMAE}.
% Experiments involving various architectures (PointMLP~\cite{pointMLP}, PointNeXt~\cite{pointNeXt}, and PointMAE~\cite{pointMAE}) are in Appendix.
For PointDA-10, we use DGCNN~\cite{dgcnn}, which serves as the backbone for most UDA works, and manually train it for each of the three datasets. As a diffusion model, we adopt $\texttt{base40M-uncond}$ from Point-E~\cite{point-e} in \href{https://github.com/openai/point-e/tree/main}{Point-E repository} and manually train it on each source dataset.% ModelNet40, ShapeNet, ModelNet, and ScanNet, respectively. %, which employs a Transformer architecture~\cite{transformer}. 
%We train $\texttt{base40M-uncond}$ on ModelNet40, ShapeNet, ModelNet, and ScanNet, respectively. 
%Details regarding the diffusion model and training hyperparameters are provided in \Cref{sec:further_implmentation_details}. %Specifically, we borrow the implementation of $\texttt{base40M-uncond}$ provided in the \href{https://github.com/openai/point-e/tree/main}{Point-E repository}.

% \vspace{-0.2in}
\subsubsection{Baselines}
To compare CloudFixer with existing TTA methods, we implement nine TTA baselines: PL~\cite{pl}, TENT~\cite{wang2020tent}, SHOT~\cite{shot}, SAR~\cite{niu2023towards}, DUA~\cite{dua}, LAME~\cite{lame}, MEMO~\cite{memo}, DDA~\cite{gao2022source}, and MATE~\cite{mirza2022mate}. The details of these baselines are in \Cref{sec:baseline_details}. They represent diverse TTA categories, including test-time training~\cite{mirza2022mate}, fully test-time adaptation~\cite{pl, wang2020tent, shot, memo, niu2023towards}, batch-norm statistics calibration~\cite{dua}, input adaptation~\cite{gao2022source}, and output adaptation~\cite{lame}. Notably, we extend DDA~\cite{gao2022source} to the 3D domain by using farthest point sampling instead of low-pass filtering with the chamfer distance for regularization. MATE~\cite{mirza2022mate} is the only test-time training method tailored for 3D point clouds. % and uniquely incorporates PointMAE \cite{pointMAE} and a masked prediction task for adaptation.
\input{tables/main_modelnet40c_bsz_1}
\input{tables/main_modelnet40c_temp_corr}
\input{tables/main_modelnet40c_imb}

\subsubsection{Implementation Details}
We conduct zero centering and scale each point cloud to a unit ball before passing it to the classifier by following~\cite{dgcnn}. Meanwhile, we standardize the point clouds to achieve a unit variance for diffusion models by following~\cite{zeng2022lion}. For diffusion models, we use the polynomial noise scheduling as in~\cite{e3_diffusion} and set the total number of timesteps as $T = 500$. The batch size~\cite{wang2020tent,shot} is set to 64 for the TTA baselines unless specified except CloudFixer and other per-sample TTA baselines (MEMO~\cite{memo}, DDA~\cite{gao2022source}) which inherently operate with a batch size of 1. % The diffusion model is trained for 5000 epochs on ModelNet40~\cite{modelnet40} with an exponential moving average (EMA) decay of 0.9999 and a batch size of 32.
For input adaptation, we perform 30 steps of updates with AdaMax optimizer~\cite{adamax}. % A learning rate linearly increases for 6 steps from 0 to 0.2 and then linearly decreases to 0.01 for the remaining steps. 
The timestep interval for the diffusion forward process is set as $[0.02T, 0.12T]$. Further details are in \Cref{sec:further_implmentation_details,sec:hyperparam_optim} and \href{https://github.com/shimazing/CloudFixer}{our repository}.%\url{https://github.com/shimazing/CloudFixer}.

\input{tables/main_modelnet40c_mild}

\subsection{Main Results}
\label{subsec:main}
\subsubsection{Results on Challenging Real-world Scenarios}
We initially assess CloudFixer in a genuinely plausible real-world scenario to verify its practical reliability. % and usability. 
We conduct validation under the batch size of 1 in~\Cref{table:main_bsz1}, considering real-time inference, temporally correlated %non-i.i.d. 
test streams in~\Cref{table:temp_correlated}, and %scenarios with 
significant label distribution shifts due to high class-imbalances in~\Cref{table:label_shift} %(we report Macro-recall only for this) 
on ModelNet40-C using Point2Vec. CloudFixer consistently outperforms other methods across various corruptions, %with 
particularly %noteworthy performance 
in noise and transformation corruptions. This results in a remarkable average performance improvement of 17\% to 19\% compared to `Unadapted' in all challenging real-world scenarios. In contrast, entropy minimization variants, such as TENT and SHOT, only achieve an accuracy of about 10\% in \Cref{table:main_bsz1} and \Cref{table:temp_correlated}, indicating model collapse. While SAR was proposed to overcome such challenging scenarios in 2D vision, it also fails in the point cloud domain. Although Per-sample adaptation methods, such as MEMO and DDA, do not collapse in these settings, MEMO shows only marginal improvement, and DDA, despite having excessive %having 4 to 5 times 
computational cost %as described in 
(see \Cref{subsubsec:efficiency}), consistently exhibits lower average performance than ours. This underscores the effectiveness of optimizing geometric transformations specialized for 3D point clouds.
% \vspace{-0.1in}

\subsubsection{Results on Mild Conditions}
While our primary focus remains on ensuring the effectiveness of our method in plausible real-world scenarios, %as mentioned in the previous paragraph, 
we also verify CloudFixer in a mild scenario of batch size of 64 and i.i.d. test stream, % with random order, 
following the standard evaluation setup used for conventional TTA methods. For this purpose, we use a Point2Vec on ModelNet40-C for common corruptions in \Cref{table:main_mild} and DGCNN on PointDA-10~\cite{qin2019pointdan} in \Cref{table:pointda} for natural distribution shift. 
In \Cref{table:main_mild}, we evaluate the online extension CloudFixer-O for comparison. 
We observe that our method consistently achieves the best performance on 9 corruptions and the second best performance on 3 corruptions out of 15, thus achieving state-of-the-art performance on average also in the mild settings. This is attributed to the effectiveness of our online model adaptation strategy, as well as our input adaptation, which can be leveraged when considering the mild conditions. A similar trend is observed with natural distribution shift in \Cref{table:pointda}.

\input{tables/main_pointda10}

\subsubsection{Across Various Classifier Architectures}
We investigate adaptation performance across different architectures to verify that CloudFixer operates in an architecture-agnostic manner due to its focus on input adaptation. Specifically, we report the average adaptation accuracy of CloudFixer across all corruptions in ModelNet40-C for various baseline architectures, including Point2Vec~\cite{point2vec}, PointMAE~\cite{pointMAE}, PointMLP~\cite{pointMLP}, and PointNeXt~\cite{pointNeXt}, along with TENT~\cite{wang2020tent}, which shows the best performance among baselines with Point2Vec, and MEMO~\cite{memo}, which is the instance-based model adaptation method. As shown in~\cref{fig:across_architecture}, we observe that CloudFixer-O achieves an average performance gain ranging from at least 13\% (PointNeXt) to a maximum of 27\% (PointMAE and PointMLP) across various corruptions, achieving the highest performance across all architectures compared to baselines. With PointMAE, only a single instance-based CloudFixer outperforms TENT with a batch size of 64. This demonstrates the architecture-agnostic benefits of CloudFixer. The detailed performance for each corruption across various architectures can be found in \Cref{subsec:architectures}.

\input{tables/ablation}

\subsection{Ablation Study}
\label{subsec:ablation}
% TODO: add visualization examples
\subsubsection{Parameterization, Objective, and Diffusion Timesteps} \label{subsubsec:ablation}
This paragraph comprehensively demonstrates whether the components of CloudFixer truly contribute to performance enhancement and if they are optimal choices. To achieve this, we validate the core components of CloudFixer, including the parameterization of geometric transformation, objective function with chamfer distance, per-point regularization, the range of timestep ($t_{min}$, $t_{max}$), the voting mechanism, and the online input adaptation in \Cref{tab:ablation}. First, we confirm the suitability of our geometric transformation parameterization of rotation and displacement. Adapting the input without parameterization (No Parameterization) or substituting the rotation matrix with a more parameter-rich affine matrix (Rotation $\rightarrow$ Affine) leads to performance degradation.
Second, we validate our optimization objective. When employing the squared $\ell_2$ loss without accounting for the unordered nature of point clouds (Squared $\ell_2$), or when naively adopting the noise matching loss of the diffusion model (Diffusion Loss), both lead to a decrease in average performance.
Next, we explore the range of timesteps and results indicate that maintaining a small value for $t$ ($t \sim U[0.01T, 0.02T]$) leads to a performance drop due to failure to overlap the source and target domains. Conversely, keeping a large value for $t$ ($t \sim U[0.4T, 0.5T]$) results in a significant drop in adaptation performance, likely due to loss of original content.
Finally, we observe that both voting (+Voting $(K=5)$) and online model adaptation (CloudFixer-O) consistently boost the performance of the original CloudFixer. Further discussions regarding the ablation study can be found in \Cref{subsec:ablation_full,subsec:ablation_distance}.
% \vspace{-0.1in}

\begin{figure}[!t]
\begin{minipage}{0.6\textwidth}
    \centering
    %\vspace{-0.1in}
    \includegraphics[width=\columnwidth]{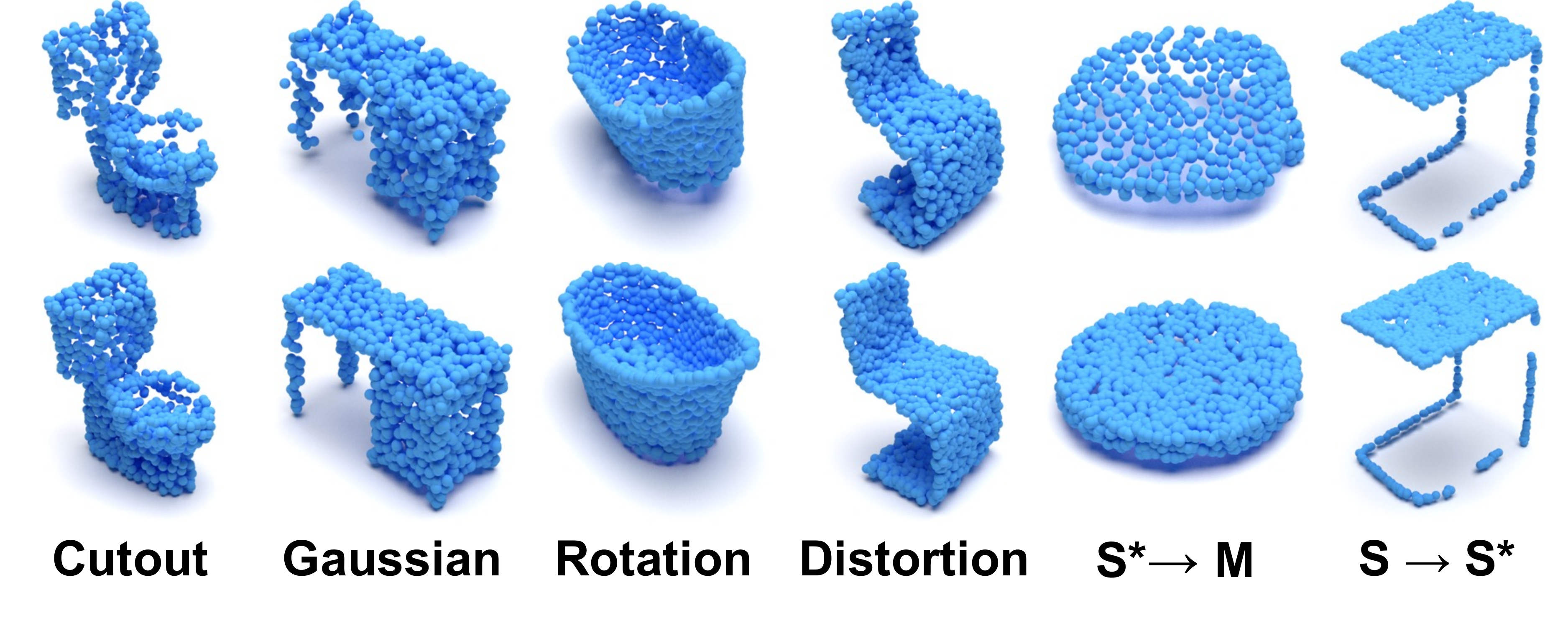}
    % \vspace{-0.16in}
    \captionof{figure}{Point cloud visualization examples demonstrate the effects of CloudFixer on various common corruption types in ModelNet40-C and natural distributions in PointDA-10. The upper row showcases corrupted examples, while the lower row illustrates the corresponding results after applying CloudFixer.}
    \label{fig:examples}
% \vspace{-.3in}
\end{minipage}
\hfill
\begin{minipage}{0.35\textwidth}
    \includegraphics[width=\columnwidth]{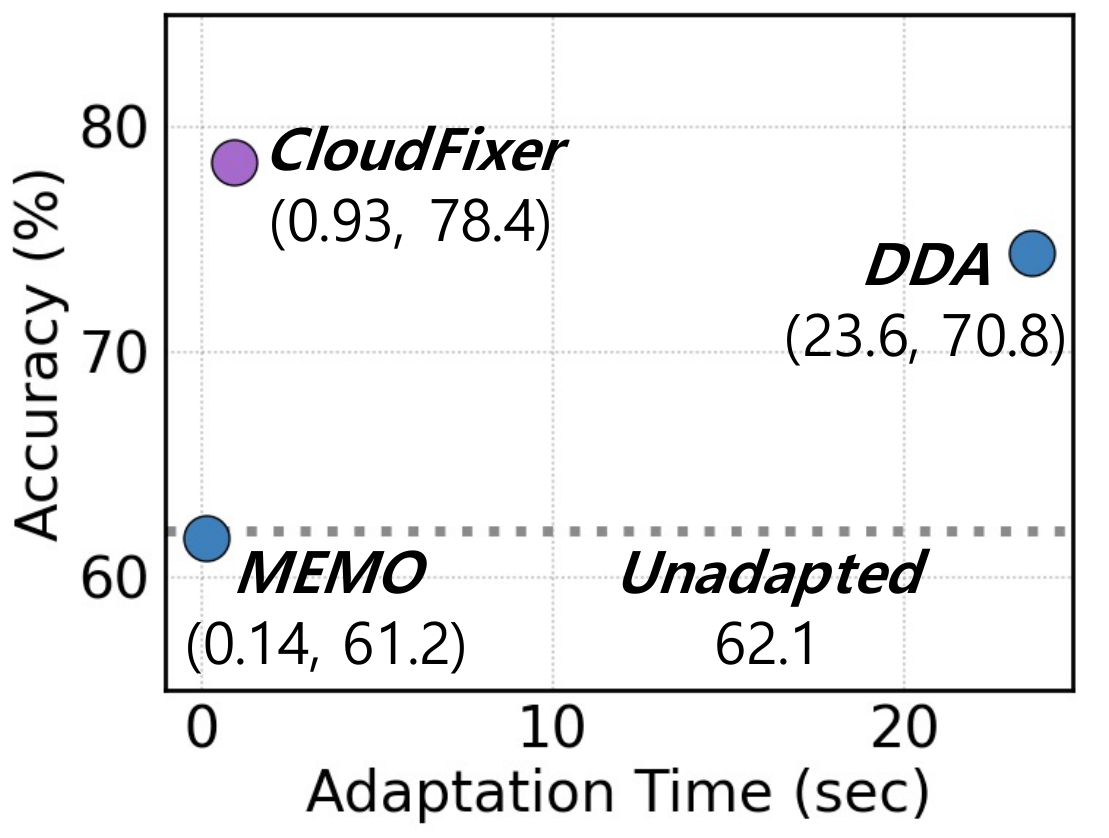}
    % \vspace{-.3in}
    \captionof{figure}{Computational cost of CloudFixer and other per-sample episodic TTA baselines by averaging the adaptation time across all corruption types within the ModelNet40-C.}
    \label{fig:efficiency}
    % \vspace{-.2in}
\end{minipage}
\end{figure}

\subsubsection{Batch Size of CloudFixer-O} \label{subsubsec:bsz}
We also evaluate the extended online model adaptation version, CloudFixer-O, compared to other baselines under varying batch sizes. \Cref{table:pointmae} is separated by a double line, with the upper part displaying results for a batch size of 1 and the lower part for a batch size of 64. To include MATE~\cite{mirza2022mate} in this comparison, we adopt the specific architecture, PointMAE~\cite{pointMAE}, utilized by MATE. `MATE($n$)' represents $n$-times model updates per batch during online adaptation. Although CloudFixer-O is designed for mild conditions, we find that it also performs well for a batch size of 1, achieving the best average accuracy even compared to the baselines with a batch size of 64. Additionally, MATE performs better than instance-based CloudFixer but worse than CloudFixer-O with a batch size of 1. Moreover, the enhancement in performance with a larger batch size is limited for MATE, in contrast to CloudFixer-O. To be specific, the performance of MATE appears to saturate at approximately 61\%, even as the number of update steps increases. On the other hand, CloudFixer-O exhibits a notable improvement, achieving an average accuracy of 72.68\% with more than a 4\% increase in accuracy points as the batch size increases from 1 to 64.
% \vspace{-.1in}

\subsection{Further Analysis}
%\vspace{-.1in}

\label{subsec:further_analysis}
\subsubsection{Qualitative Analysis}
\label{subsubsec:qualitative}
In this section, we assess whether the performance improvement observed in CloudFixer is incidental or can be attributed to the actual transformation of shifted test instances into the source domain through CloudFixer's input adaptation strategy. To investigate this, we visually examine point clouds before and after adaptation across different target domains, as illustrated in \cref{fig:examples}. Additionally, we present more adaptation examples for all common corruptions in \cref{fig:examples1} to \cref{fig:examples3}. As depicted in \cref{fig:examples}, we observe that CloudFixer successfully restores corrupted target instances for each common corruption---Cutout, Gaussian, Rotation, and Distortion---back to clean instances from the source domain of ModelNet40. It is worth noting that this phenomenon extends to natural distribution shifts as well; irregular point clouds from ScanNet are transformed into regular patterns resembling those from ModelNet (S$^* \rightarrow$ M) , while conversely, the legs of a desk from ShapeNet are converted into the irregular form of ScanNet (S $\rightarrow$ S$^*$). We also provide more adaptation examples in \Cref{sec:examples}.
% \vspace{-.1in}

\subsubsection{Computational Efficiency}
\label{subsubsec:efficiency}
% One might wonder about the real-time feasibility of CloudFixer, which employs iterative optimization using a diffusion model.
To assess the real-time feasibility of CloudFixer, we conduct a comparison with per-sample episodic TTA baselines (MEMO, DDA). The assessment involves averaging the adaptation time across all corruption types within the ModelNet40-C, utilizing a single RTX 3090 GPU. CloudFixer %, leveraging a pre-trained diffusion model, 
achieves an acceptable adaptation speed within one second (0.93 seconds on average) with a moderate number of iterations and the absence of backpropagation through the diffusion model or extensive iterative generation. This stands in stark contrast to DDA, which also utilizes the same diffusion model architecture as CloudFixer but takes approximately 23.6 seconds, making it prohibitively costly for practical real-time usage. Moreover, CloudFixer can take the advantage of batch processing. Combining this observation with our hyperparameter sensitivity experiments, we anticipate further reduction in adaptation time by employing a number of iterations smaller than 30.
% \vspace{-.1in}

\begin{figure}[!t]
    \centering
    % \vspace{-.2in}
    \includegraphics[width=.9\textwidth]{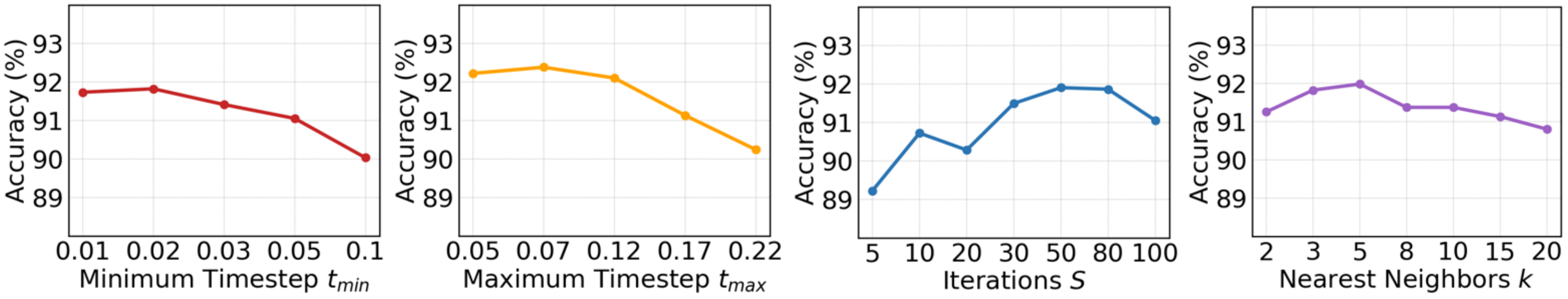}
    % \vspace{-.25in}
    \caption{Hyperparameter sensitivity analysis regarding $(t_{min}$, $t_{max})$ for the forward process, the number of adaptation steps $S$, and the number of nearest neighbors $k$ used in displacement regularization using Point2Vec on IP of ModelNet40-C.}
    \label{fig:hparam_sensitivity}
    % \vspace{-.25in}
\end{figure}
% \vspace{-.1in}

\subsubsection{Hyperparameter Sensitivity}
To evaluate hyperparameter sensitivity, we conduct an analysis involving the minimum and maximum timesteps $t_{min}$, $t_{max}$ for the forward process, the number of adaptation steps $S$, and the number of nearest neighbors $k$ used in displacement regularization using Point2Vec on `IP' of ModelNet40-C. While keeping the remaining hyperparameters fixed as best with $(t_{min}, t_{max}, S, k) = (0.02, 0.12, 30, 5)$, we vary one at a time. Considering the unadapted performance at 66.25\%, as depicted in \cref{fig:hparam_sensitivity}, CloudFixer illustrates the establishment of a sweet spot for all critical hyperparameters, indicating its insensitivity to hyperparameters. This highlights a significant advantage in utilization during inference time, where hyperparameter optimization is infeasible due to uncertainty about potential distribution shifts.
\label{subsubsec:hparam_sensitivity}

% \vspace{-.1in}

%% file: tables/main_modelnet40c_bsz_1.tex
\begin{table*}[t]
    \centering
    \caption{Accuracy on ModelNet40-C with a limited batch size of 1 using Point2Vec.}
    \label{table:main_bsz1}
    % \vspace{-.15in}
    \setlength{\tabcolsep}{4pt}
    \renewcommand{\arraystretch}{1.3}
    \resizebox{0.9\linewidth}{!}{
    \begin{tabular}{cl|ccccc|ccccc|ccccc|c}

    \Xhline{3\arrayrulewidth}

    & \multirow{2}{*}{\bf Method} & \multicolumn{5}{c|}{\bf Density Corruptions} &  \multicolumn{5}{c|}{\bf Noise Corruptions} &  \multicolumn{5}{c|}{\bf Transformation Corruptions} & \multirow{2}{*}{\bf Avg.} \\ \cmidrule(lr){3-7} \cmidrule(lr){8-12} \cmidrule(lr){13-17}
    
    & & \bf OC & \bf LD & \bf DI & \bf DD & \bf CO & \bf UNI & \bf GAU & \bf IP & \bf US & \bf BG & \bf ROT & \bf SH & \bf FFD & \bf RBF & \bf IR & \\ \Xhline{2\arrayrulewidth}

    & Unadapted & 41.09 & 20.02 & 90.03 & \underline{86.51} & 83.91 & 63.41 & 49.68 & 66.25 & 38.33 & 37.68 & 47.04 & 79.29 & 75.97 & 74.68 & 77.51 & 62.09 \\ \cdashline{2-18}
    
    \multirow{5}{*}{\rotatebox{90}{\textbf{Online}}} & PL~\cite{pl} & 30.11 & 4.38 & \bf 91.65 & \bf 88.82 & \bf 85.74 & 88.78 & 5.88 & 80.83 & 4.70 & 5.02 & 30.92 & 83.71 & 81.32 & 82.17 & 83.91 & 56.53 \\
    
    & TENT~\cite{wang2020tent} & 4.05 & 4.05 & 5.27 & 4.29 & 5.55 & 4.38 & 4.05 & 3.89 & 4.05 & 4.05 & 4.05 & 4.17 & 4.05 & 4.05 & 4.42 & 4.29 \\
    
    & SHOT~\cite{shot} & 4.01 & 4.05 & 4.05 & 4.05 & 4.09 & 4.05 & 4.05 & 4.01 & 4.01 & 4.05 & 4.05 & 4.05 & 4.05 & 4.05 & 4.05 & 4.04 \\

    & SAR~\cite{niu2023towards} & 16.37 & 5.71 & 90.44 & 83.43 & 83.47 & 81.40 & 73.10 & 32.17 & 14.83 & 24.03 & 19.73 & 77.51 & 75.93 & 76.54 & 78.04 & 55.51 \\
    
    & DUA~\cite{dua} & 41.37 & 20.26 & 89.71 & 86.10 & 83.71 & 62.56 & 49.64 & 65.68 & 37.48 & 36.95 & 47.37 & \bf 79.58 & 75.57 & 73.70 & 77.23 & 61.79 \\ \cline{2-18}
    
    \multirow{3}{*}{\rotatebox{90}{\textbf{Episodic}}} & LAME~\cite{lame} & 40.96 & 20.50 & 90.19 & 85.70 & 84.16 & 63.57 & 49.92 & 66.25 & 38.33 & 37.32 & 47.04 & 79.34 & 75.24 & 74.72 & 77.59 & 62.06 \\
    
    & MEMO~\cite{memo} & \bf 46.27 & 36.67 & \underline{90.76} & 82.94 & 81.77 & 79.90 & 76.18 & 80.88 & 70.91 & 32.94 & 63.57 & \underline{79.74} & \underline{79.21} & 78.28 & 81.16 & 70.75 \\
    
    & DDA~\cite{gao2022source} & \underline{42.67} & 36.30 & 89.26 & 83.10 & \underline{84.44} & 90.44 & 89.38 & \underline{90.76} & 86.26 & 53.69 & 49.68 & 78.69 & 77.43 & 80.71 & 82.78 & 74.37 \\ \cline{2-18}
    
    & \bf CloudFixer & 41.00 & \underline{38.82} & 87.32 & 80.27 & 83.06 & \underline{91.09} & \underline{90.52} & \underline{90.76} & \underline{89.06} & \underline{75.49} & \underline{81.04} & 78.28 & 78.73 & \underline{82.98} & \underline{85.09} & \underline{78.23} \\
    & \bf + Voting ($K=5$) & 41.00 & \bf 38.90 & 88.05 & 80.55 & 84.16 & \bf 91.45 & \bf 91.29 & \bf 91.98 & \bf 89.79 & \bf 76.58 & \bf 83.51 & 79.38 & \bf 79.70 & \bf 84.12 & \bf 85.90 & \bf{79.09} \\
    
    \Xhline{3\arrayrulewidth}
\end{tabular}}
% \vspace{-.1in}
\end{table*}

%% file: tables/main_modelnet40c_temp_corr.tex
\begin{table*}[t]
    \centering
    \caption{Accuracy on ModelNet40-C with temporally correlated non-i.i.d. test stream using Point2Vec, where the test set is sorted based on label order.}
    \label{table:temp_correlated}
    % \vspace{-.15in}
    \setlength{\tabcolsep}{4pt}
    \renewcommand{\arraystretch}{1.3}
    \resizebox{0.9\linewidth}{!}{
    \begin{tabular}{cl|ccccc|ccccc|ccccc|c}

    \Xhline{3\arrayrulewidth}
    
    & \multirow{2}{*}{\bf Method} & \multicolumn{5}{c|}{\bf Density Corruptions} &  \multicolumn{5}{c|}{\bf Noise Corruptions} &  \multicolumn{5}{c|}{\bf Transformation Corruptions} & \multirow{2}{*}{\bf Avg.} \\ \cmidrule(lr){3-7} \cmidrule(lr){8-12} \cmidrule(lr){13-17}
    
    & & \bf OC & \bf LD & \bf DI & \bf DD & \bf CO & \bf UNI & \bf GAU & \bf IP & \bf US & \bf BG & \bf ROT & \bf SH & \bf FFD & \bf RBF & \bf IR & \\ \Xhline{2\arrayrulewidth}

    & Unadapted & 41.33 & 14.22 & 90.15 & 75.20 & 75.93 & 63.01 & 50.32 & \underline{67.06} & 37.60 & 36.75 & 47.61 & \underline{79.05} & 75.45 & 74.68 & 77.39 & 60.38 \\ \cdashline{2-18}

    \multirow{5}{*}{\rotatebox{90}{\textbf{Online}}} & PL~\cite{pl} & 16.69 & 11.43 & 24.03 & 21.88 & 20.14 & 20.83 & 19.77 & 20.10 & 18.64 & 9.76 & 16.90 & 22.45 & 20.75 & 21.03 & 20.62 & 19.00 \\

    & TENT~\cite{wang2020tent} & 12.16 & 8.59 & 16.61 & 16.33 & 14.26 & 14.55 & 14.75 & 14.91 & 14.55 & 6.24 & 8.35 & 10.25 & 11.51 & 15.92 & 13.01 & 12.80 \\

    & SHOT~\cite{shot} & 12.16 & 8.59 & 16.61 & 16.33 & 14.26 & 14.55 & 14.75 & 14.91 & 14.55 & 6.24 & 8.35 & 10.25 & 11.51 & 15.92 & 13.01 & 12.80 \\

    & SAR~\cite{niu2023towards} & 16.61 & 10.86 & 23.78 & 20.79 & 20.58 & 20.79 & 19.00 & 18.68 & 17.79 & 9.85 & 17.26 & 22.29 & 21.07 & 20.79 & 20.75 & 18.73 \\

    & DUA~\cite{dua} & 19.29 & 13.49 & 26.86 & 24.96 & 23.74 & 24.11 & 22.53 & 23.10 & 21.23 & 12.12 & 19.21 & 24.68 & 23.62 & 23.70 & 22.49 & 21.68 \\ \cline{2-18}

    \multirow{3}{*}{\rotatebox{90}{\textbf{Episodic}}} & LAME~\cite{lame} & \bf 43.52 & 14.14 & \bf 95.38 & \underline{84.44} & \bf 84.76 & 70.46 & 53.12 & 73.22 & 40.92 & 41.45 & \underline{51.34} & \bf 86.14 & \bf 83.75 & \underline{81.48} & \underline{84.64} & 65.92 \\

    & MEMO~\cite{memo} & 40.76 & 20.38 & \underline{90.28} & \bf 85.90 & 84.00 & 62.40 & 49.23 & 65.68 & 37.52 & 36.67 & 47.16 & \underline{79.05} & 75.32 & 74.43 & 77.47 & 61.75 \\

    & DDA~\cite{gao2022source} & \underline{42.67} & \underline{36.30} & 89.26 & 83.10 & \underline{84.44} & \underline{90.44} & \underline{89.38} & \bf 90.76 & \underline{86.26} & \underline{53.69} & 49.68 & 78.69 & 77.43 & 80.71 & 82.78 & \underline{74.37} \\ \cline{2-18}

    & \bf CloudFixer & 41.00 & \bf 38.82 & 87.32 & 80.27 & 83.06 & \bf 91.09 & \bf 90.52 & \bf 90.76 & \bf 89.06 & \bf 75.49 & \bf 81.04 & 78.28 & \underline{78.73} & \bf 82.98 & \bf 85.09 & \bf 78.23 \\

    \Xhline{3\arrayrulewidth}
    \end{tabular}}
    % \vspace{-0.2in}
\end{table*}

%% file: tables/main_modelnet40c_imb.tex
\begin{table*}[t]
    \centering
    \caption{Macro-recall on ModelNet40-C with label distribution shifts, featuring a high class imbalance ratio of 100, using Point2Vec.}
    \label{table:label_shift}
    % \vspace{-.15in}
    \setlength{\tabcolsep}{4pt}
    \renewcommand{\arraystretch}{1.3}
    \resizebox{0.9\linewidth}{!}{
    \begin{tabular}{cl|ccccc|ccccc|ccccc|c}

    \Xhline{3\arrayrulewidth}
    
    & \multirow{2}{*}{\bf Method} & \multicolumn{5}{c|}{\bf Density Corruptions} &  \multicolumn{5}{c|}{\bf Noise Corruptions} &  \multicolumn{5}{c|}{\bf Transformation Corruptions} & \multirow{2}{*}{\bf Avg.} \\ \cmidrule(lr){3-7} \cmidrule(lr){8-12} \cmidrule(lr){13-17}
    
    & & \bf OC & \bf LD & \bf DI & \bf DD & \bf CO & \bf UNI & \bf GAU & \bf IP & \bf US & \bf BG & \bf ROT & \bf SH & \bf FFD & \bf RBF & \bf IR & \\ \Xhline{2\arrayrulewidth}

    & Unadapted & 47.59 & 21.29 & \bf 88.78 & \underline{82.03} & 79.76 & 57.77 & 46.04 & 59.41 & 35.81 & 29.25 & 49.88 & \underline{75.81} & 75.62 & 75.95 & 74.07 & 59.94 \\ \cdashline{2-18}

    \multirow{5}{*}{\rotatebox{90}{\textbf{Online}}} & PL~\cite{pl} & 47.87 & 39.53 & 83.04 & 80.33 & 77.18 & 74.62 & 68.10 & 69.55 & 65.95 & 28.03 & 56.35 & 73.57 & 72.15 & 75.28 & 75.79 & 65.82 \\

    & TENT~\cite{wang2020tent} & 35.71 & 32.99 & 76.93 & 74.12 & 65.13 & 64.46 & 66.74 & 63.95 & 66.09 & 27.39 & 46.49 & 55.69 & 65.42 & 69.19 & 67.04 & 58.49 \\

    & SHOT~\cite{shot} & 35.44 & 28.90 & 66.72 & 65.08 & 64.71 & 56.88 & 59.36 & 58.21 & 58.92 & 33.41 & 37.10 & 61.73 & 64.25 & 62.40 & 66.92 & 54.67 \\

    & SAR~\cite{niu2023towards} & \bf 49.07 & 40.27 & 85.63 & 79.98 & 75.48 & 71.33 & 68.28 & 68.08 & 64.04 & 28.58 & 55.48 & 72.34 & 75.44 & 75.73 & 72.84 & 65.50 \\

    & DUA~\cite{dua} & 51.03 & \bf 42.56 & 85.16 & \bf 82.89 & 81.98 & 77.39 & 71.37 & 74.81 & 67.55 & 29.78 & \underline{59.86} & \bf 78.07 & \underline{76.95} & \underline{76.25} & 77.45 & 68.87 \\ \cline{2-18}

    \multirow{3}{*}{\rotatebox{90}{\textbf{Episodic}}} & LAME~\cite{lame} & 45.37 & 15.91 & 86.52 & 75.81 & 72.24 & 49.42 & 40.09 & 51.46 & 27.22 & 25.15 & 43.77 & 67.49 & 68.43 & 65.64 & 67.46 & 53.47 \\

    & MEMO~\cite{memo} & \underline{47.97} & 19.53 & \underline{88.31} & 81.67 & 80.26 & 57.00 & 47.05 & 57.19 & 34.01 & 28.54 & 49.26 & 72.84 & 73.63 & 74.36 & 75.28 & 59.13 \\

    & DDA~\cite{gao2022source} & 47.88 & 39.49 & 84.93 & 80.77 & \underline{82.58} & \underline{84.71} & \underline{85.77} & \underline{85.49} & \underline{78.81} & \underline{43.07} & 51.67 & 74.79 & 75.23 & 75.55 & \underline{78.79} & \underline{71.30} \\ \cline{2-18}

    & \bf CloudFixer & 44.43 & \underline{40.91} & 86.87 & 77.23 & \bf 83.75 & \bf 91.46 & \bf 91.72 & \bf 92.23 & \bf 85.80 & \bf 75.53 & \bf 76.83 & 75.36 & \bf 78.68 & \bf 80.11 & \bf 83.04 & \bf 77.60 \\

    \Xhline{3\arrayrulewidth}
    \end{tabular}}
\end{table*}

%% file: tables/main_modelnet40c_mild.tex
\begin{table*}[t]
    \centering
    \caption{Accuracy on ModelNet40-C using Point2Vec under the mild conditions of a batch size of 64 and an i.i.d. test stream.}
    \label{table:main_mild}
    % \vspace{-.15in}
    \setlength{\tabcolsep}{4pt}
    \renewcommand{\arraystretch}{1.3}
    \resizebox{0.9\linewidth}{!}{
    \begin{tabular}{cl|ccccc|ccccc|ccccc|c}

    \Xhline{3\arrayrulewidth}

    & \multirow{2}{*}{\bf Method} & \multicolumn{5}{c|}{\bf Density Corruptions} &  \multicolumn{5}{c|}{\bf Noise Corruptions} &  \multicolumn{5}{c|}{\bf Transformation Corruptions} & \multirow{2}{*}{\bf Avg.} \\ \cmidrule(lr){3-7} \cmidrule(lr){8-12} \cmidrule(lr){13-17}
    
    & & \bf OC & \bf LD & \bf DI & \bf DD & \bf CO & \bf UNI & \bf GAU & \bf IP & \bf US & \bf BG & \bf ROT & \bf SH & \bf FFD & \bf RBF & \bf IR & \\ \Xhline{2\arrayrulewidth}

    & Unadapted & 41.09 & 20.02 & 90.03 & \underline{86.51} & 83.91 & 63.41 & 49.68 & 66.25 & 38.33 & 37.68 & 47.04 & 79.29 & 75.97 & 74.68 & 77.51 & 62.09 \\ \cdashline{2-18}
    
    & PL~\cite{pl} & 46.07 & 38.17 & 90.56 & 84.20 & 82.13 & 83.06 & 78.81 & 81.77 & 76.66 & 38.05 & 63.49 & 81.40 & 79.09 & 79.21 & 81.36 & 72.27 \\
    
    & TENT~\cite{wang2020tent} & \underline{47.33} & 37.88 & \underline{91.25} & \bf 87.44 & \underline{86.43} & \underline{89.14} & \underline{87.80} & \underline{85.41} & \underline{87.64} & \underline{67.91} & 71.23 & \bf 85.37 & \bf 85.05 & \bf 86.30 & \underline{86.51} & \underline{78.85} \\
    
    & SHOT~\cite{shot} & 46.72 & \bf 47.45 & 86.83 & 84.60 & 82.90 & 80.92 & 79.78 & 74.92 & 80.92 & 61.51 & \underline{74.31} & 78.48 & 81.16 & 81.73 & 82.54 & 74.98 \\
    
    & SAR~\cite{niu2023towards} & 46.27 & 36.67 & 90.76 & 82.94 & 81.77 & 79.90 & 76.18 & 80.88 & 70.91 & 32.94 & 63.57 & 79.74 & 79.21 & 78.28 & 81.16 & 70.75 \\
    
    & DUA~\cite{dua} & \bf 47.85 & 39.55 & \bf 91.45 & 85.25 & 83.51 & 81.28 & 77.84 & 82.62 & 74.92 & 38.49 & 66.00 & 82.66 & 80.23 & \underline{80.51} & 82.94 & 73.01 \\
    
    & LAME~\cite{lame} & 39.18 & 9.52 & 90.03 & 76.30 & 76.54 & 61.79 & 46.31 & 63.90 & 35.09 & 31.93 & 45.75 & 78.44 & 75.04 & 74.35 & 76.74 & 58.73 \\ \cline{2-18}
    
    %& MEMO~\cite{memo} & 46.27 & 36.67 & 90.76 & 82.94 & 81.77 & 79.90 & 76.18 & 80.88 & 70.91 & 32.94 & 63.57 & 79.74 & 79.21 & 78.28 & 81.16 & 70.75 \\
    
    %\rot{\rlap{\textbf{Episodic}}} & DDA~\cite{gao2022source} & 42.67 & 36.30 & 89.26 & 83.10 & 84.44 & 90.44 & 89.38 & 90.76 & 86.26 & 53.69 & 49.68 & 78.69 & 77.43 & 80.71 & 82.78 & 74.37 \\ \cline{2-18}
    
    %& \bf CloudFixer & 41.00 & 38.82 & 87.32 & 80.27 & 83.06 & 91.09 & 90.52 & 90.76 & 89.06 & 75.49 & 81.04 & 78.28 & 78.73 & 82.98 & 85.09 & 78.23 \\
    
    %& \bf + Voting (K=3) & 41.00 & 38.90 & 88.05 & 80.55 & 84.16 & 91.45 & 91.29 & 91.98 & 89.79 & 76.58 & 83.51 & 79.38 & 79.70 & 84.12 & 85.90 & 79.09 \\
    
    & \bf \footnotesize CloudFixer-O & 46.39 & \underline{44.94} & 90.92 & 84.76 & \bf 86.99 & \bf  91.94 & \bf 91.86 & \bf 91.82 & \bf 92.14 & \bf 74.92 & \bf 85.98 & \underline{83.83} & \underline{82.09} & \bf 86.30 & \bf 87.40 & \bf 81.49 \\

    \Xhline{3\arrayrulewidth}
    
    \end{tabular}}
% \vspace{-0.2in}
\end{table*}

%% file: tables/main_pointda10.tex
%\begin{table*}[ht]
\begin{figure}[!t]
\begin{minipage}{0.48\textwidth}
\centering
\captionof{table}{Accuracy on PointDA-10 using DGCNN. We report the performance of all method, except MEMO, DDA, CloudFixer, including CloudFixer-O with a batch size of 64. % We use $K=3$ for voting of CloudFixer.
}
\label{table:pointda}
% \vspace{-0.15in}
\setlength{\tabcolsep}{2pt}
\renewcommand{\arraystretch}{1.3}
\resizebox{\linewidth}{!}{
\begin{tabular}{cl|cccccc|c}
%\hline
\Xhline{3\arrayrulewidth}
& \bf Method & \bf{M $\rightarrow$ S} & \bf{M $\rightarrow$ S*} & \bf{S $\rightarrow$ M} & \bf{S $\rightarrow$ S*} & \bf{S* $\rightarrow$ M} & \bf{S* $\rightarrow$ S} & \bf Avg. \\ \Xhline{2\arrayrulewidth}
    & Unadapted & 81.38 & 52.23 & 77.10 & 44.83 & 62.03 & 64.69 & 63.71 \\ \cdashline{2-9}

    \multirow{5}{*}{\rotatebox{90}{\textbf{Online}}} & PL~\cite{pl} & 75.52 & 50.99 & 66.82 & \underline{50.42} & 58.88 & 61.12 & 60.63 \\

    & TENT~\cite{wang2020tent} & 76.77 & 52.63 & 65.07 & 49.46 & 57.59 & 59.31 & 60.14 \\
    
    & SHOT~\cite{shot} & 14.53 & 14.53 & 67.17 & 47.77 & 44.63 & 19.62 & 34.71 \\
    
    & SAR~\cite{niu2023towards}& 74.12 & 49.75 & 66.36 & 50.03 & 54.32 & 58.83 & 58.90 \\
    
    & DUA~\cite{dua} & 74.80 & 50.25 & 67.29 & 49.24 & 59.93 & 60.03 & 60.26 \\ \cline{2-9}
    
    \multirow{3}{*}{\rotatebox{90}{\textbf{Episodic}}} & LAME~\cite{lame} & 82.26 & \bf 53.93 & \bf 80.84 & 44.83 & 20.68 & 43.42 & 54.33 \\ 
    
    & MEMO~\cite{memo} & 75.84 & 47.26 & 73.25 & 50.99 & 47.90 & 60.27 & 59.25 \\
    
    & DDA~\cite{gao2022source} & 82.46 & 51.89 & \underline{80.72} & 44.77 & \underline{65.42} & 66.85 & \underline{65.35} \\ \cline{2-9}
    
    \multirow{3}{*}{\rotatebox{90}{\textbf{Ours}}} & \textbf{CloudFixer} & \underline{83.46} & 53.12 & 77.80 & 43.30 & 63.97 & 67.85 & 64.92 \\
    
    & \textbf{+ Voting ($K=3$)} & \bf 83.51 & \underline{53.47} & 77.92 & 43.64 & 64.48 & \underline{68.37} & 65.23 \\
    
    &\bf CloudFixer-O &  80.42 &  53.08 & 75.00 & \bf 51.55 & \bf 66.00 & \bf 69.02 & \bf 65.85 \\
\Xhline{3\arrayrulewidth}
\end{tabular}}
% \vspace{-.2in}
\end{minipage}
% \hfill
\hspace{.03in}
\begin{minipage}{0.48\textwidth}
    \centering
    \includegraphics[width=\columnwidth]{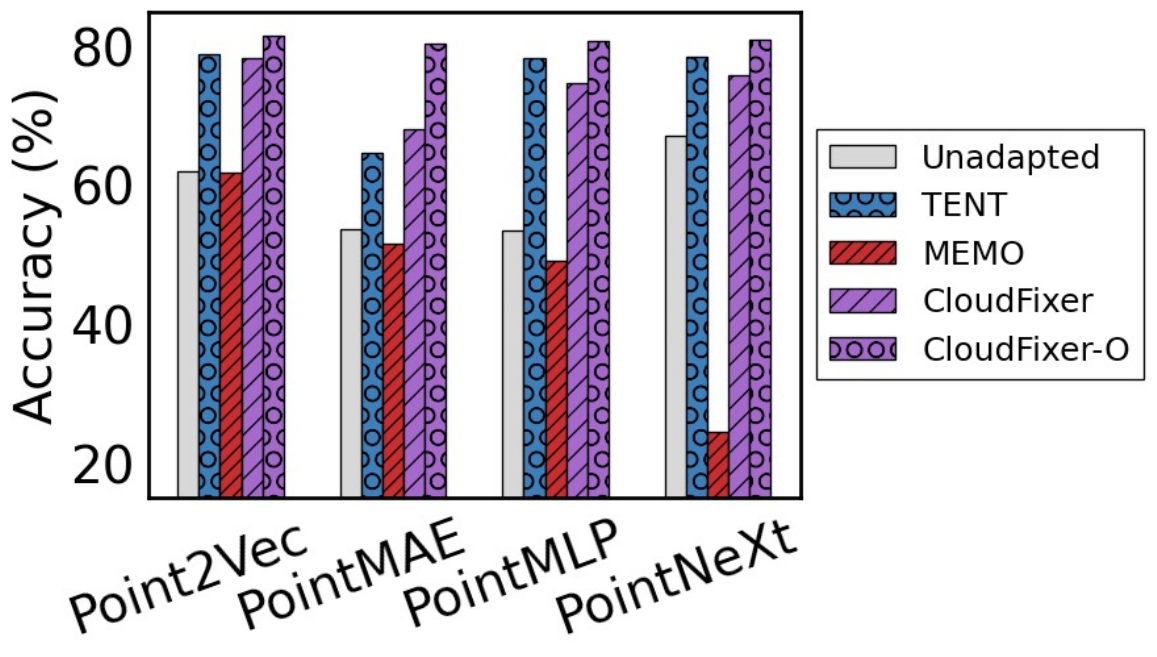}
     % \vspace{-.2in}
    \captionof{figure}{
    The average accuracy of CloudFixer and other baselines across all corruptions in ModelNet40-C for various classifier architectures.
    \label{fig:across_architecture}
    % \vspace{-.2in}
    }
\end{minipage}
\end{figure}

%% file: tables/ablation.tex
%\begin{table}[ht]
\begin{figure}[!t]
\begin{minipage}{.53\textwidth}
\centering
\captionof{table}{Ablation study of the core strategies in CloudFixer, including parameterization, objective, displacement regularization, forward timesteps, voting, and online adaptation.}
%\vspace{-.15in}
%\resizebox{0.8\columnwidth}{!}{
\resizebox{\textwidth}{!}{
\setlength{\tabcolsep}{4pt}
\renewcommand{\arraystretch}{1.4}
\label{tab:ablation}
\begin{tabular}{l|cccccc|c}

\Xhline{3\arrayrulewidth}

\multirow{2}{*}{\bf Setting} & \multicolumn{2}{c|}{\bf Density} & \multicolumn{2}{c|}{\bf Noise} & \multicolumn{2}{c|}{\bf Transform} & \multirow{2}{*}{\bf Avg.} \\ \cmidrule(lr){2-3} \cmidrule(lr){4-5} \cmidrule(lr){6-7}

& \bf LD & \bf CO & \bf US & \bf BG & \bf ROT & \bf RBF & \\ \Xhline{2\arrayrulewidth}

Unadapted & 20.02 & 83.91 & 38.33 & 37.68 & 47.04 & 74.68 & 51.30 \\ \hdashline

No Parameterization & 39.02 & 82.13 & \underline{91.81} & 75.64 & 49.79 & 80.87 & 70.85 \\
Rotation $\rightarrow$ Affine & 37.88 & 82.50 & 87.16 & 74.96 & 66.05 & 82.66 & 72.59 \\ \hline

Squared $\ell_2$ & 37.12 & 83.71 & 71.80 & \underline{76.34} & 78.93 & 80.23 & 72.47 \\
Diffusion Loss & 33.14 & 73.70 & 86.95 & 39.10 & 59.08 & 70.46 & 62.18 \\ \hline

No Reg. & 36.26 & 73.26 & 85.17 & 61.30 & 55.71 & 79.05 & 65.27 \\
Uniform Reg. & \underline{39.30} & 89.10 & 84.68 & 48.82 & 80.02 & 83.87 & 71.49 \\ \hline

$t \sim U[0.01T, 0.02T]$ & 21.03 & 82.86 & 84.93 & 54.09 & 58.79 & 40.07 & 58.10 \\
$t \sim U[0.4T, 0.5T]$ & 17.06 & 48.10 & 38.61 & 58.14 & 60.01 & 37.16 & 42.72 \\ \hline

\bf CloudFixer & 38.82 & 83.06 & 89.06 & 75.49 & 81.04 & 82.98 & 75.79 \\
\bf +Voting ($K=5$) & 38.90 & \underline{84.16} & 89.79 & \bf 76.58 & \underline{83.51} & \underline{84.12} & \underline{76.83} \\
\bf CloudFixer-O & \bf 44.94 & \bf 86.99 & \bf 92.14 & 74.92 & \bf 85.98 & \bf 86.30 & \bf 79.20 \\

\Xhline{3\arrayrulewidth}

\end{tabular}
}
% \vspace{-0.1in}
\end{minipage}
% \hfill
% \vspace{-0.1in}
\hspace{.03in}
\begin{minipage}{0.422\textwidth}
    \centering
    \captionof{table}{Accuracy on ModelNet40-C, using TTA baselines including MATE~\cite{mirza2022mate} with batch size of 1 and 64 using PointMAE.}
    \label{table:pointmae}
    \resizebox{\textwidth}{!}{
    \setlength{\tabcolsep}{4pt}
    \renewcommand{\arraystretch}{1.4}
    
    \begin{tabular}{cl|cccccc|c}
       \Xhline{3\arrayrulewidth}
        &\textbf{Method} & \textbf{OC} & \textbf{LD} & \textbf{DD} & \textbf{GAU} & \textbf{BG} & \textbf{IR} & \textbf{Avg.} \\
        \Xhline{2\arrayrulewidth}

        & Unadapted & 34.72 & 15.19 & 78.65 & 51.30 & 16.87 & 62.03 & 43.13 \\ \cdashline{1-9}
        
        \multirow{5}{*}{\rotatebox{90}{\textbf{Bsz. 1}}} &MEMO~\cite{memo} & 33.18 & 14.79 & 78.08 & 47.57 & 6.81 & 59.97 & 40.07 \\
        
        & DDA~\cite{gao2022source} & 39.91 & 37.07 & 79.70 & \underline{87.20} & 19.89 & 76.30 & 56.68 \\ \cline{2-9}
        
        & MATE(1)~\cite{mirza2022mate} & \bf 52.55 & \bf 49.85 & \bf 82.54 & 78.97 & 13.05 & \underline{78.44} & \underline{59.23} \\ \cline{2-9}
        
        & \bf CloudFixer & 34.24 & 35.62 & 71.15 & 87.16 & \underline{50.36} & 72.45 & 58.50 \\
        & \bf CloudFixer-O & \underline{47.53} & \underline{48.95} & \underline{81.52} & \bf 88.98 & \bf 59.44 & \bf 84.76 & \bf 68.53 \\ \hline \noalign{\vskip\doublerulesep \vskip-\arrayrulewidth} \hline

        \multirow{10}{*}{\rotatebox{90}{\textbf{Bsz. 64}}}  &PL~\cite{pl} & 48.01 & 41.45 & 79.13 & 70.38 & 17.83 & 70.18 & 54.50 \\
        & TENT~\cite{wang2020tent} & 47.53 & 41.82 & 79.94 & 73.38 & 17.30 & 71.27 & 55.21 \\
        & SHOT~\cite{shot} & \bf 55.71 & \underline{50.16} & 72.33 & 67.34 & 14.59 & 67.75 & 54.65 \\
        & SAR~\cite{niu2023towards} & 47.73 & 45.62 & 80.79 & 72.00 & 10.98 & 72.20 & 54.89 \\
        & DUA~\cite{dua} & 50.85 & 45.83 & 83.14 & 75.24 & 19.17 & 75.41 & 58.27 \\
        & LAME~\cite{lame} & 34.89 & 16.05 & 80.92 & 50.49 & 8.51 & 62.40 & 42.21 \\ \cline{2-9}

        & MATE(1)~\cite{mirza2022mate} & 48.52 & 36.51 & \underline{84.40} & 71.07 & 11.18 & 72.49 & 54.03 \\
        & MATE(10)~\cite{mirza2022mate} & 54.09 & 46.47 & 84.28 & 76.34 & 19.69 & 78.97 & 59.97 \\
        & MATE(20)~\cite{mirza2022mate} & \underline{55.06} & 48.87 & 83.31 & \underline{76.42} & \underline{21.35} & \underline{80.19} & \underline{60.87} \\\cline{2-9}
        
        & \bf CloudFixer-O & 52.55 & \bf 53.61 & \bf 84.76 & \bf 90.56 & \bf 67.50 & \bf 87.07 & \bf 72.68 \\
        \Xhline{3\arrayrulewidth}
    \end{tabular}}
% \vspace{-0.1in}
\end{minipage}
\end{figure}

% \begin{table}[h]
% \centering
% \caption{Ablation study of \emph{CloudFixer} \B{with respect to parametrization, forward timestep, and regularization.}}
% \vspace{-.15in}
% \resizebox{0.7\columnwidth}{!}{
% \setlength{\tabcolsep}{4pt}
% \renewcommand{\arraystretch}{1.4}
% \label{tab:ablation}
% \begin{tabular}{l|ccccccc|c}

%     \Xhline{3\arrayrulewidth}

%     \bf Setting & \bf LD & \bf DI & \bf UNI & \bf GAU & \bf US & \bf RBF & \bf IR & \bf Avg. \\ \Xhline{2\arrayrulewidth}
 
%     Unadapted & 25.81 & 90.32 & 86.02 & 84.08 & 81.40 & 84.24 & 86.14 & 76.86 \\ \hdashline

%     \B{$t = 0$} & \B{4.38} & \B{80.35} & \B{86.26} & 84.00 & 81.20 & & 85.62 & \\
    
%     $t \sim [0.6T, T]$ & 8.31 & 51.34 & 54.38 & 42.30 & 44.25 & 54.74 & 62.36 & 45.38 \\

%     Affine. & 25.61 & 89.18 & 86.43 & 84.28 & 80.79 & 84.12 & 85.41 & 76.55 \\

%     No Reg. & 36.75 & 79.38 & 87.28 & 88.29 & 82.33 & 82.09 & 83.51 & 77.09 \\ \hline
    
%     \bf CloudFixer & \bf 41.28 & \bf 89.38 & \bf 89.71 & \bf 89.87 & \bf 85.98 & \bf 86.12 & \bf 86.91 & \bf 81.32 \\
    
%     \Xhline{3\arrayrulewidth}
% \end{tabular}}
% \vspace{-.18in}
% \end{table}

%% file: sections/conclusion.tex
\section{Conclusion}
In this paper, we have introduced CloudFixer, a novel test-time input adaptation method for 3D point clouds. Leveraging a pre-trained diffusion model's domain translation capability, CloudFixer directly optimizes carefully designed geometric transformation parameters to translate input point clouds, taking into account computational costs. Extensive experiments show that our method achieves state-of-the-art results across diverse distribution shift scenarios. Our approach advances test-time input adaptation for 3D point cloud recognition, highlighting essential design principles that integrate geometric optimization and diffusion model knowledge. 

%% file: sections/acknowledgement.tex
\section*{Acknowledgments}
This work was supported by the Institute for Information \& Communications Technology Planning \& Evaluation (IITP) grant funded by the Korea government (MSIP) (No.2019-0-00075, Artificial Intelligence Graduate School Program (KAIST)).

%% file: sections/appendix.tex
\clearpage
\appendix
\counterwithin{figure}{section}
\counterwithin{table}{section}
\renewcommand\thefigure{\thesection\arabic{figure}}
\renewcommand\thetable{\thesection\arabic{table}}

\centerline{\Large\bf Appendix}

\section{Algorithm of CloudFixer} \label{sec:alg}
We present the detailed algorithm of CloudFixer in \Cref{alg:cloudfixer}, which encompasses both the original input adaptation and the online model adaptation. It is worth noting that the pre-trained diffusion model $\epsilon_\theta$ is frozen and utilized solely to acquire $\hat{y}$, guiding $y$ back to the source domain.
\input{algorithms/cloudfixer.tex}

\section{Originality and Advantages over DDA} \label{sec:orig_over_dda}
While DDA~\cite{gao2022source} and CloudFixer both use diffusion models for input adaptation, their core ideas and methods are distinct. DDA uses an iterative generative process while preserving low-frequency information, whereas CloudFixer uses a parameterized 3D geometric transformation guided by the source diffusion model. Consequently, \emph{DDA struggles with misalignment and structural corruption}, leading to \emph{significant performance differences}, especially in BG and ROT. DDA also requires more iterative steps and backpropagation through diffusion models, making it \emph{over 25 times slower than CloudFixer} (\cref{fig:efficiency}).
Last but not least, the online model adaptation utilizing the consistency proposed by us is a distinctive feature of our model, clearly distinguished by DDA.

\section{Limitations and Broader Impacts}
\subsubsection{Limitations}
One prevalent test-time corruption for 3D point clouds is occlusion, where the input data is incomplete. When the source domain contains clean point clouds and the target domain involves occluded point clouds, such domain translation problem becomes a point cloud completion task. Since the pre-trained diffusion model operates on normalized point clouds, and normalizing severely occluded point clouds to be zero-centered with a unit standard deviation results in a significant scale and center location shift for clean point clouds, this presents a complex research problem. For example, in the case of an occluded chair with only the backrest visible, the backrest's scale may increase, shifting downward and potentially causing misclassification as a monitor. Achieving accurate translation for complete chairs requires substantial point movement. However, since CloudFixer regularizes large steps, our model currently has limitations in addressing this specific corruption type. Developing TTA methods to effectively handle such corruption stands as a promising future research direction.

\subsubsection{Broader Impacts}
The increasing demand for computer vision, particularly in 3D vision applications like autonomous systems and virtual reality, has stimulated significant research into domain adaptation methods to address distribution shifts. Test-time adaptation holds the potential to enhance the performance of 3D perception models in autonomous systems, encompassing applications such as self-driving cars, drones, and robotics. In this regard, the application of our method allows for the translation of test data into a source domain format without the need for source data in on-device settings, thereby conducting input adaptation. This capability is poised to have a significant impact on the evolving field of 3D vision, fostering further growth in the field. We hope this research provides valuable insights to the academic community and enhances the robustness of point cloud recognition models in real-world scenarios.

\section{Reproducibility Statement}
For reproducibility, we provide our implementation code in our \href{https://github.com/shimazing/CloudFixer/tree/main}{github repository}.
%\small{$\texttt{cloudfixer/}$}. 
%Unfortunately, due to file size constraints, we are unable to include the pre-trained diffusion model trained on ModelNet40 
%\noindent\small{$\texttt{cloudfixer/ckpt/diffusion\_model\_transformer\_modelnet40.npy}$} at this time. However, we plan to make it available to the public at a later date.  % and pre-trained diffusion models \small{$\texttt{cloudfixer/ckpt/diffusion\_model\_transformer\_modelnet40.npy}$}
%in the directory named \small{$\texttt{cloudfixer/ckpt/}$} as a supplementary material. 
We supply the bash scripts \small{$\texttt{scripts/train\_dm.sh}$} for training a diffusion model on ModelNet40 and 
\small{$\texttt{scripts/run\_cloudfixer.sh}$} for adaptation on ModelNet40-C using CloudFixer and \small{$\texttt{scripts/run\_baselines.sh}$} for adaptation on ModelNet40-C using baselines, respectively. Our hyperparameter optimization results for TTA baselines, which are elucidated in~\Cref{sec:hyperparam_optim} can be found in \small{$\texttt{cfgs/hparams}$}. For a pair comparison and to ensure reproducibility, we consistently set the random seed to 2 for all conducted experiments. %Additionally, we plan to make our source code publicly available. 
All hyperparameters and associated details are in our source code.
Lastly, checkpoints for ModelNet40 classifiers can be downloaded in the following links for each architecture: \href{https://github.com/kabouzeid/point2vec/tree/17f8ad80b78017f9fc74986a5b20f453abfca9b5}{Point2Vec}, \href{https://github.com/ma-xu/pointMLP-pytorch/tree/main}{PointMLP}, \href{https://github.com/guochengqian/PointNeXt/tree/master}{PointNeXt}, and \href{https://github.com/jmiemirza/MATE}{PointMAE}. We refer readers to \texttt{README.md} for the remaining details.

\section{Dataset Details}
\label{sec:dataset_details}
\subsection{ModelNet40-C}
\begin{figure*}[t]
  \centering
  \includegraphics[width=\textwidth]{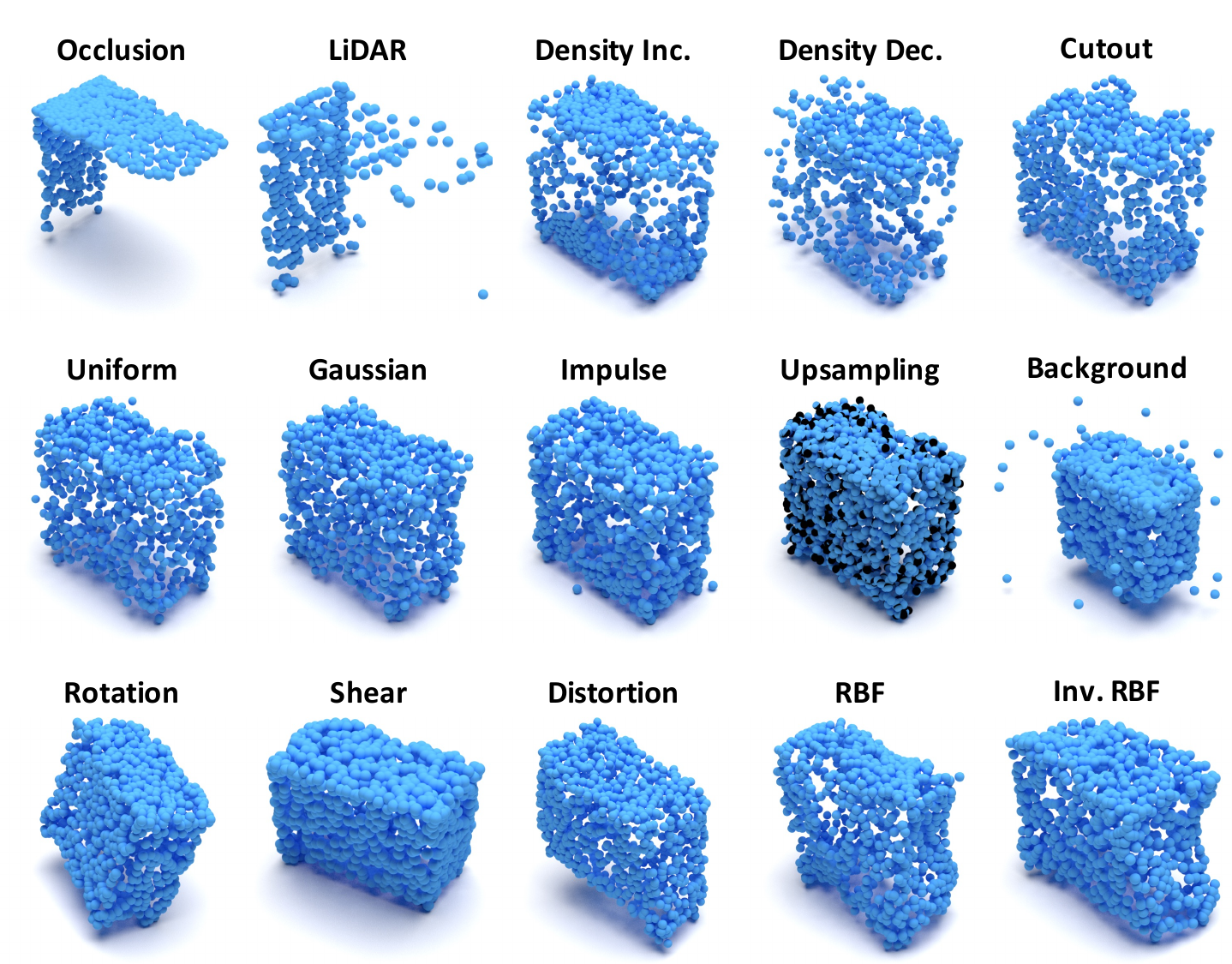}
  \caption{An illustrative example depicting 15 distinct corruption types---Occlusion, LiDAR, Density Inc., Density Dec., Cutout, Uniform, Gaussian, Impulse, Upsampling, Background, Rotation, Shear, Distortion, RBF, Inv. RBF---for a single original point cloud associated with the dresser class in the test set of ModelNet40-C.}
  \label{fig:modelnet40c_examples}
\end{figure*}

\cref{fig:modelnet40c_examples} showcases 15 corruptions of \href{https://github.com/jiachens/ModelNet40-C}{ModelNet40-C}~\cite{modelnet40c} on which we conduct the experiments. ModelNet40-C involves applying synthetic corruptions to the test set of ModelNet40 which comprises 2468 point clouds of 40 classes. ModelNet40-C contains broad corruption types categorized into three classes---density corruptions, noise corruptions, and transformation corruptions. For all experiments, we use severity level 5. We refer the readers to the \href{https://github.com/jiachens/ModelNet40-C}{official repository}.

\subsection{PointDA-10}
\begin{figure*}[t]
  \centering
  \includegraphics[width=\textwidth]{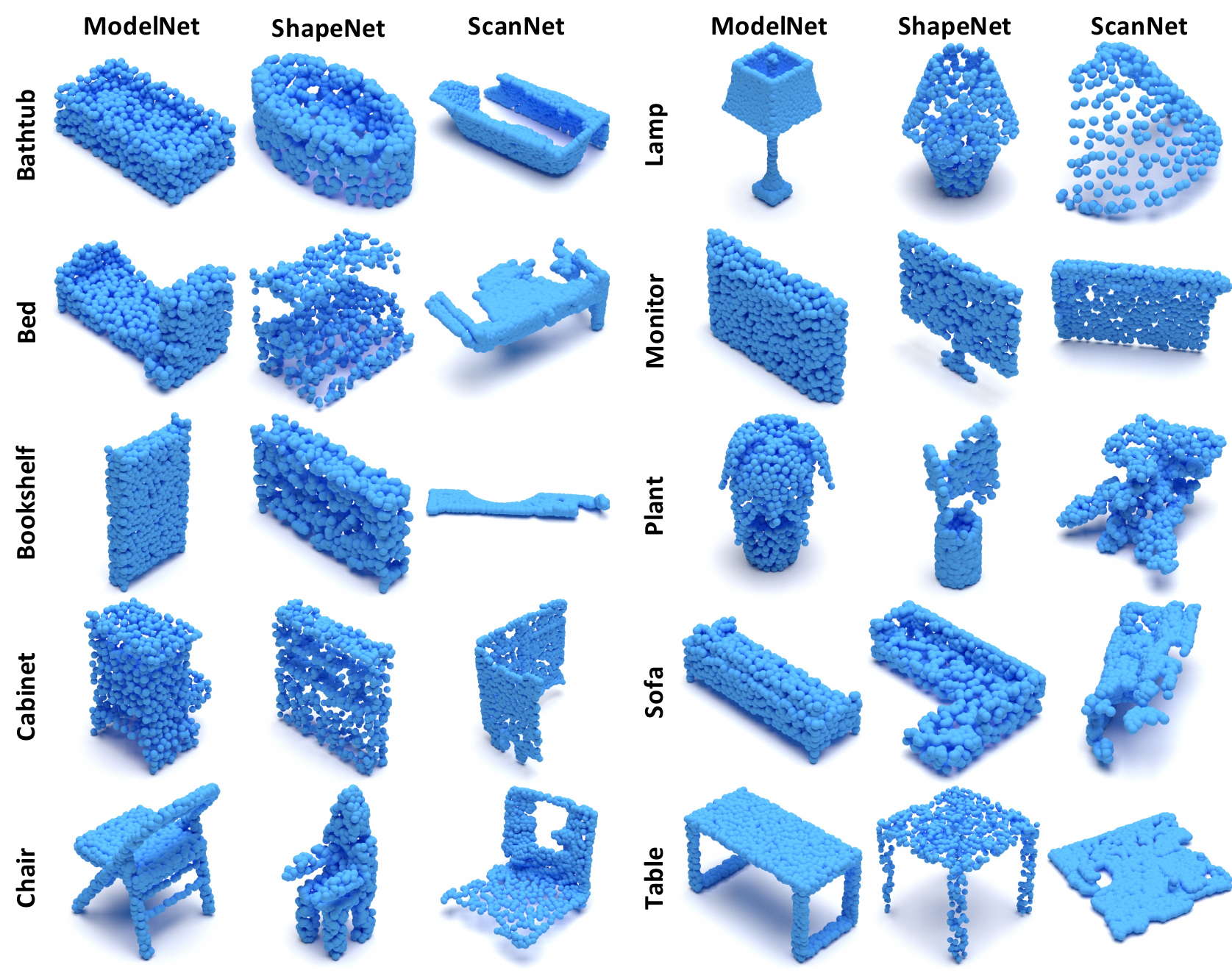}
  \caption{An illustrative example showcasing 10 different classes---Bathtub, Bed, Bookshelf, Cabinet, Chair, Lamp, Monitor, Plant, Sofa, Table---in PointDA-10.}
  \label{fig:pointda}
\end{figure*}

PointDA-10~\cite{qin2019pointdan} incorporates natural distribution shifts in real-world scenarios. This dataset is originally proposed to serve as a benchmark for unsupervised domain adaptation by using two out of the three datasets---ModelNet~\cite{modelnet}, ShapeNet~\cite{shapenet}, and ScanNet~\cite{scannet}---as the source and target domains. The benchmark is composed of 10 common classes shared across the three datasets. ModelNet and ShapeNet are synthetically generated from 3D CAD models, while ScanNet is created by scanning real-world scenes. Their class distributions are not even but imbalanced. Examples are provided in \cref{fig:pointda}.

\section{Baseline Details}
\label{sec:baseline_details}
\subsubsection{PL}
Pseudo-Labeling (PL)~\cite{pl} conducts pseudo-labeling based on the model's predictions and updates the model weights at test-time through cross-entropy loss. PL involves two crucial test-time hyperparameters: learning rate and adaptation steps.

\subsubsection{TENT}
Test ENTropy minimization (TENT)~\cite{wang2020tent} is a test-time adaptation method in which the model undergoes fine-tuning using an unsupervised loss with the objective of reducing the prediction entropy of the model. TENT entails two important test-time hyperparameters: learning rate and adaptation steps.

\subsubsection{SHOT}
Source HypOthesis Transfer (SHOT)~\cite{shot} is a fully test-time adaptation method that employs unsupervised objectives such as entropy minimization, diversity maximization, and self-supervised pseudo-labeling. It also updates the statistics of batch normalization layers using the statistics of the test batch. SHOT incorporates three critical test-time hyperparameters: learning rate, adaptation steps, and the pseudo-labeling loss weight, to adjust the relative weight of the loss induced by self-supervised pseudo-labeling.

\subsubsection{SAR}
Sharpness-Aware and Reliable entropy minimization (SAR)~\cite{niu2023towards} utilizes entropy minimization, but it excludes instances with high prediction entropy to prevent model collapse and incorporates the Sharpness-Aware Minimization (SAM)~\cite{sam}, fostering the model's adaptation to flat minima. SAR has four essential test-time hyperparameters: learning rate, adaptation steps, an entropy threshold to exclude adaptation from high entropy samples, and an epsilon threshold employed in the calculation of sharpness in SAM~\cite{sam}.

\subsubsection{DUA}
Dynamic Unsupervised Adaptation (DUA)~\cite{dua} is a test-time adaptation method that focuses on calibrating batch normalization statistics. At the onset of test time, all batch normalization layers are set to a trained state. For each incoming test batch, the model undergoes multiple forward passes, iteratively updating batch normalization statistics as moving averages. DUA utilizes two test-time hyperparameters: adaptation steps and a decay factor, which dictates the rate at which the moving average progresses.

\subsubsection{LAME}
Laplacian Adjusted Maximum-likelihood Estimation (LAME)~\cite{lame} operates by updating the model's prediction output rather than the model parameters. It conducts on a batch of given test data, adjusting the probabilities of similar samples in the feature space to be similar when the original prediction probabilities for the test data batch are provided.  LAME is characterized by three crucial test-time hyperparameters: the kernel affinity, used to define the kernel density function, the number of neighbors to be regarded as similar, and the maximum number of steps for iterative output probability optimization.

\subsubsection{MEMO}
Marginal Entropy Minimization with One test point (MEMO)~\cite{memo} is a single-instance TTA technique. It utilizes multiple data augmentations to generate multiple predictions for the test instance. Subsequently, it minimizes the entropy of the average probability of these predictions. This approach not only achieves the effect of traditional entropy minimization but also encourages similarity among predictions from different viewpoints. MEMO is defined by three key test-time hyperparameters: learning rate, adaptation steps, and the number of augmentations that can be applied to each test instance.

\subsubsection{DDA}
Diffusion-Driven Adaptation (DDA)~\cite{gao2022source} is a single-instance test-time adaptation for 2D image classification tasks using a diffusion model, similar to our approach. DDA is characterized by two essential test-time hyperparameters: guidance weight, which determines the extent to which the original content is preserved, and the low-pass filtering scale. Unlike our method, DDA is a generation-based method that employs low-pass filtering in the reverse process to preserve class information after the forward process. In contrast to this approach, our method achieves significantly faster execution times by avoiding back-propagation through the diffusion model and delivers superior performance by leveraging optimization-based techniques tailored for 3D tasks.

\subsubsection{MATE}
MATE~\cite{mirza2022mate} is a test-time training approach for 3D point clouds, enhancing deep network robustness in point cloud classification against distribution shifts. It employs a masked autoencoder test-time objective, constructing batches of 48 from single corrupted point clouds, randomly masking 90\% of each sample, and fine-tuning the model using reconstruction loss. MATE efficiently adapts with minimal fractions of points from each test sample, sometimes as low as 5\%, making it lightweight for real-time applications. However, MATE necessitates manipulation of the training procedure and relies on the specific architecture of PointMAE for self-reconstruction.

\section{Further Implementation Details}
\label{sec:further_implmentation_details}
% preprocessing
\subsubsection{Pre-processing} Before passing point clouds to classifiers, we conduct zero centering and scale it to a unit ball, adhering to the original configuration outlined in~\cite{dgcnn} for each point cloud. In our pre-processing step for inputs of diffusion models, following zero centering, we standardize the point clouds to achieve a unit variance following~\cite{zeng2022lion}.

% diffusion pretraining
\subsubsection{Pre-training Diffusion Models} For 4 source datasets---ModelNet40-C, ShapeNet, ModelNet, and ScanNet---we use the same setting as follows. We use the polynomial noise scheduling used in~\cite{e3_diffusion} and set the total number of timesteps of the diffusion model as $T = 500$. The diffusion model is trained for 5000 epochs on each dataset with an exponential moving average (EMA) decay of 0.9999 and a batch size of 64.

% baselines
\subsubsection{Baselines} For model adaptation methods (PL, TENT, SHOT, SAR, and MEMO), we also update batch normalization statistics by initializing all layers to a trained state at the onset of the inference phase. PL, TENT, and SAR exclusively fine-tune the affine parameters of batch normalization statistics following the default settings of previous works~\cite{wang2020tent, niu2023towards}, while SHOT and MEMO optimize all parameters. For PL, TENT, SHOT, and SAR, we configure all settings to be online, where the model is not reset to the pre-trained state upon receiving the test batch. In contrast, for LAME, MEMO, and DDA, we set all settings to be episodic, where the model is reset to the pre-trained state every time a test batch is received. We adopt a standard batch size of 64 for the online TTA baseline method, following common practice~\cite{wang2020tent, shot}, unless otherwise specified. Notably, our proposed method, CloudFixer, along with per-sample TTA baselines (MEMO, DDA), can operate with a batch size of 1.

% CloudFixer
\subsubsection{CloudFixer}
With regard to our input adaptation method, we employ a consistent configuration across all datasets. We conduct 30 iterations of updates utilizing the AdaMax optimizer~\cite{adamax} with a learning rate that linearly increases for 6 steps (20\% of warmup for total steps) from 0 to 0.2 and then linearly decreases to 0.01 for the remaining steps. 
The timestep interval for the diffusion forward process is defined as $[t_{min}, t_{max}]$ with $t_{min}=0.02T$ and $t_{max}=0.12T$. For the computation of weights ${w_j}_j$ to regulate $\delta_j$, we designate the number of nearest neighbors as $k=5$. For regularization, $\lambda(\cdot)$ is initialized to 10 and cosine annealed to 1 for the 30 steps.

% CloudFixer-O
\subsubsection{CloudFixer-O}
Under mild conditions, we propose to adapt model parameters as well as inputs. Model adaptation is conducted in an online manner. With a batch size of 64, we update models one time per batch using AdamW with a learning rate $10^{-5}$ and $10^{-4}$ on ModelNet40-C and PointDA-10, respectively. For each instance, we obtain 3 different transformations. Furthermore, we report the results of CloudFixer-O with a batch size of 1 on ModelNet40-C using PointMAE. In this case, we update models one time per batch using AdamW with a learning rate $10^{-6}$. We use a smaller learning rate because the update step increases as a batch size decreases. For each instance, we obtain 48 different transformations following MATE \cite{mirza2022mate}.

\section{Hyperparameter Optimization} \label{sec:hyperparam_optim}
Due to the well-known sensitivity of hyperparameters in TTA methods~\cite{gao2022source, niu2023towards}, we conduct rigorous hyperparameter optimization when reproducing the baselines. As optimizing hyperparameters for each test dataset is unfeasible in real-world scenarios, we tune the hyperparameters of each TTA method using the original test set of the source classifier. Generally, the batch size is a critical hyperparameter in test-time adaptation, and while increasing it typically improves adaptation performance, it cannot be increased indefinitely given the online nature of the practical test-time situations. Therefore, we initially set the batch size to 64 following the standard setting~\cite{niu2023towards, gao2022source}. Subsequently, we optimize the hyperparameters of each method by conducting a random search with a maximum iteration of 30. Our hyperparameter search space is reported in \Cref{table:tta_hyperparam_searchspace}.

\input{tables/hparam_space.tex}

\section{Additional Experiments}
\subsection{Results for All Corruption Types Across Various Architectures}
\label{subsec:architectures}
In this subsection, we present the adaptation results under mild conditions, employing a batch size of 64 and i.i.d. test stream, which is an extension of the findings depicted in \cref{fig:across_architecture}. We consider three additional architectures: PointMLP~\cite{pointMLP} in \Cref{table:pointmlp}, PointNeXt in \Cref{table:pointnext}, and PointMAE~\cite{pointMAE} in \Cref{table:pointmae_full}. Note here that we maintain the same test-time hyperparameters used in the Point2Vec experiments. We find that CloudFixer consistently shows its superiority regardless of architecture. Moreover, CloudFixer-O also brings successful performance enhancement across diverse classifier architectures.

\input{tables/pointMLP}
\input{tables/pointNeXt}
\input{tables/pointMAE}

\subsection{Ablation Study on All Corruption Types}
\label{subsec:ablation_full}
\input{tables/ablation_full}
In our main paper, we partially report the ablation studies through \Cref{tab:ablation} due to space constraints. We also show the ablation results on all corruptions of ModelNet40C in \Cref{table:ablation_full}. As in \Cref{tab:ablation}, the originally proposed setting, CloudFixer, achieves the best performance on average even when considering all corruptions. Similar to the ablation study in the main paper \Cref{subsec:ablation}, this consistently validates the significant performance improvement contributed by various components of CloudFixer, including geometric transformation parameterization, objective function with chamfer distance, per-point regularization, timestep range ($t_{min}$, $t_{max}$), voting mechanism, and online input adaptation, affirming their optimality. However, we can observe that ablated settings demonstrate improved performance in certain cases. For example, in the case of Shear (SH) corruption, utilizing affine transformation instead of rotation yields the best performance. This observation can be attributed to the original definition of Shear transformation which falls within affine transformations.

\subsection{Further Ablation Study on Distance Metrics}
\label{subsec:ablation_distance}
\begin{figure*}[t]
  \centering
  \includegraphics[width=\textwidth]{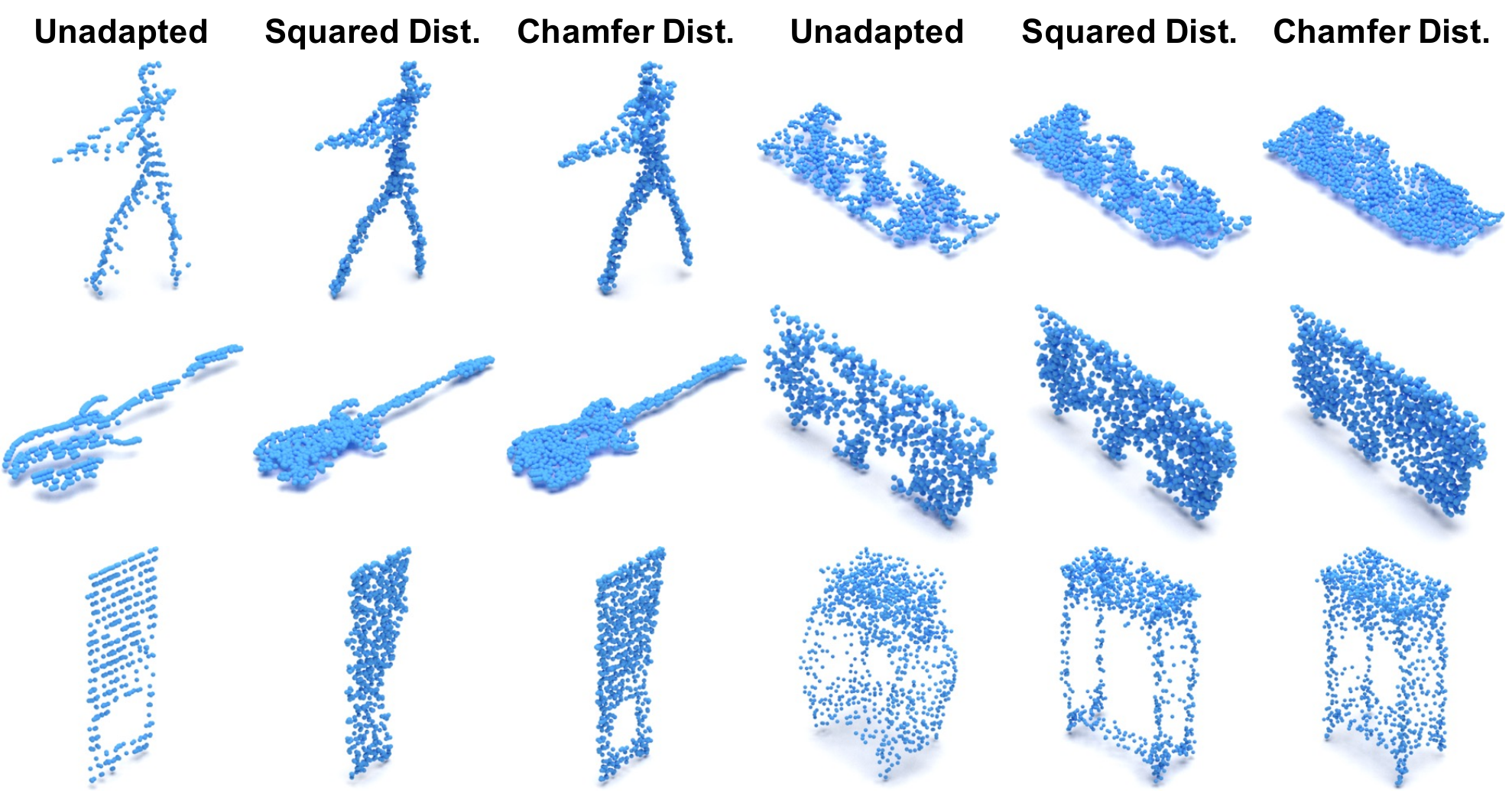}
  \caption{Qualitative analysis of the ablation study on distance metrics with illustrative examples.}
  \label{fig:ablation_chamfer}
\end{figure*}

We propose using Chamfer distance as our distance metric between a point cloud undergoing adaptation and an estimate from the diffusion model. This choice is motivated by the unordered nature of the points in point clouds. The other possible choice is to naively use squared $\ell_2$ distance which is used to train our diffusion model.
%However, \Cref{tab:ablation_chamfer} showcases that Chamfer distance consistently outperforms the other option. This result indicates considering the nature of point clouds is important for stable adaptation.
We present various adaptation examples in \cref{fig:ablation_chamfer}, using two different distance metrics. % The illustration showcases that Chamfer distance consistently outperforms the other option.
The adapted results with Chamfer distance consistently appear to be clearer than the others. This indicates considering the nature of point clouds is important for stable adaptation.

\subsection{Comparison with Data Augmentation}
In this section, we compare CloudFixer with data augmentation strategies, which are commonly used in domain generalization. Domain generalization methods like MetaSets~\cite{huang2021metasets} might be effective for specific shifts like sim-to-real transfer, but they heavily rely on presumed augmentations, limiting efficacy in TTA scenarios with arbitrary target domains. Besides, CloudFixer operates at test time, complementing rather than competing with train time domain generalization. MetaSets’ augmentations improve Point2Vec on ModelNet40-C (Avg.) from 62.09\% to 68.75\%, notably enhancing ‘Density Corruptions’ (\eg, OC: 41.09\% to 61.46\%), but show no improvement for ‘Transformation Corruptions.’ CloudFixer-O further improves the performance to \textbf{83.05\%}, handling corruptions MetaSets could not, showing domain generalization and ours are complementary.

\input{tables/metasets}

\subsection{Efficacy on Adversarial Attack}
Another potential application of CloudFixer is its effectiveness in countering adversarial attacks. To this end, we compare the adaptation performance for adversarial examples generated using the projected gradient descent method~\cite{pgd} on an original clean test set of ModelNet40 using PointMLP in~\Cref{tab:adv}. The adversarial attack utilized a step size of $4 \cdot 10^{-3}$ over 30 steps, with a bound of 0.16 for each coordinate, where both the step size and the bound are measured as $\ell_{\infty}$ distance. We could compare with adversarial robustness-focused methods, but they often require specialized training. Since our primary focus is on TTA, we opt to compare with TTA methods. Some TTA methods such as TENT, SAR, and LAME fail to recover performance significantly. However, CloudFixer demonstrates substantially higher performance compared to other benchmarks, achieving accuracy levels close to the oracle. This underscores the reliability of CloudFixer in test scenarios where natural distribution shifts may occur or unintended attacks may arise.

\input{tables/adv_attack}

\section{Adaptation Examples}
\label{sec:examples}
\begin{figure*}[t]
  \centering
  \includegraphics[width=\textwidth]{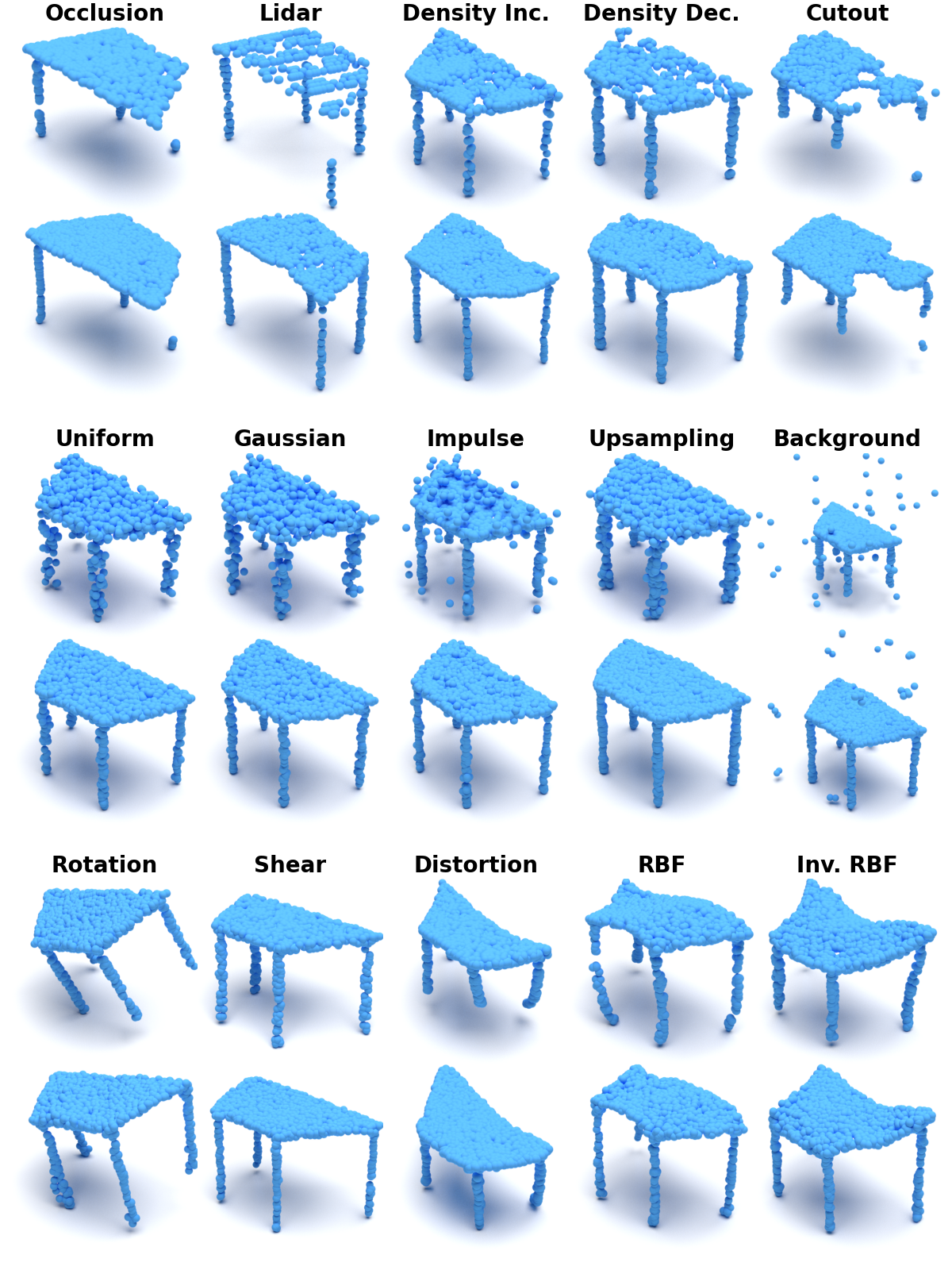}
  \caption{Comparison between the corrupted and adapted point clouds following the application of CloudFixer to 15 different corruption types. The example pertains to a single original point cloud belonging to the desk class in the test set of ModelNet40-C.}
  \label{fig:examples1}
\end{figure*}

\begin{figure*}[t]
  \centering
  \includegraphics[width=\textwidth]{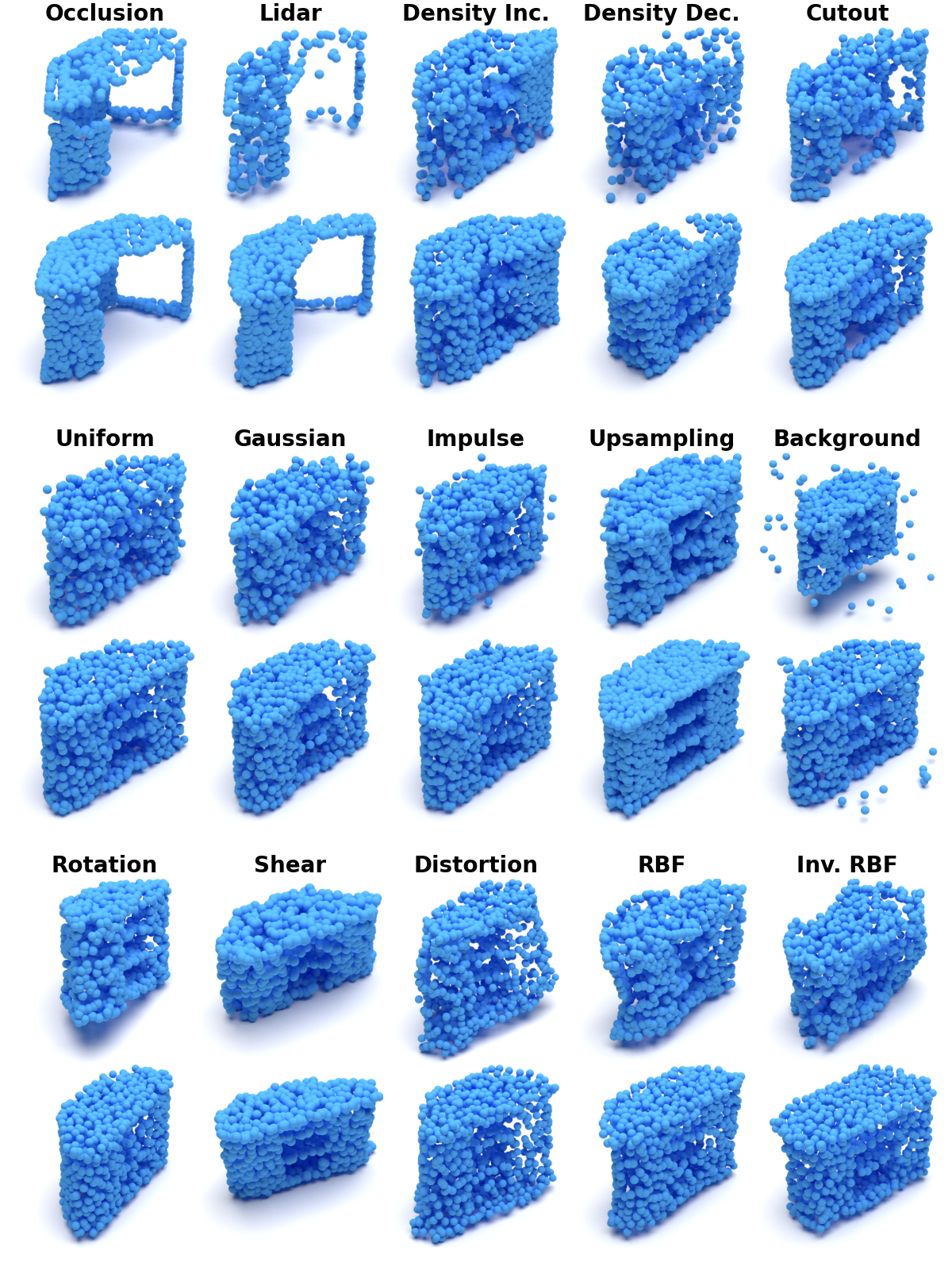}
  \caption{Comparison between the corrupted and adapted point clouds following the application of CloudFixer to 15 different corruption types. The example pertains to a single original point cloud belonging to the table class in the test set of ModelNet40-C.}
  \label{fig:examples2}
\end{figure*}

\begin{figure*}[t]
  \centering
  \includegraphics[width=\textwidth]{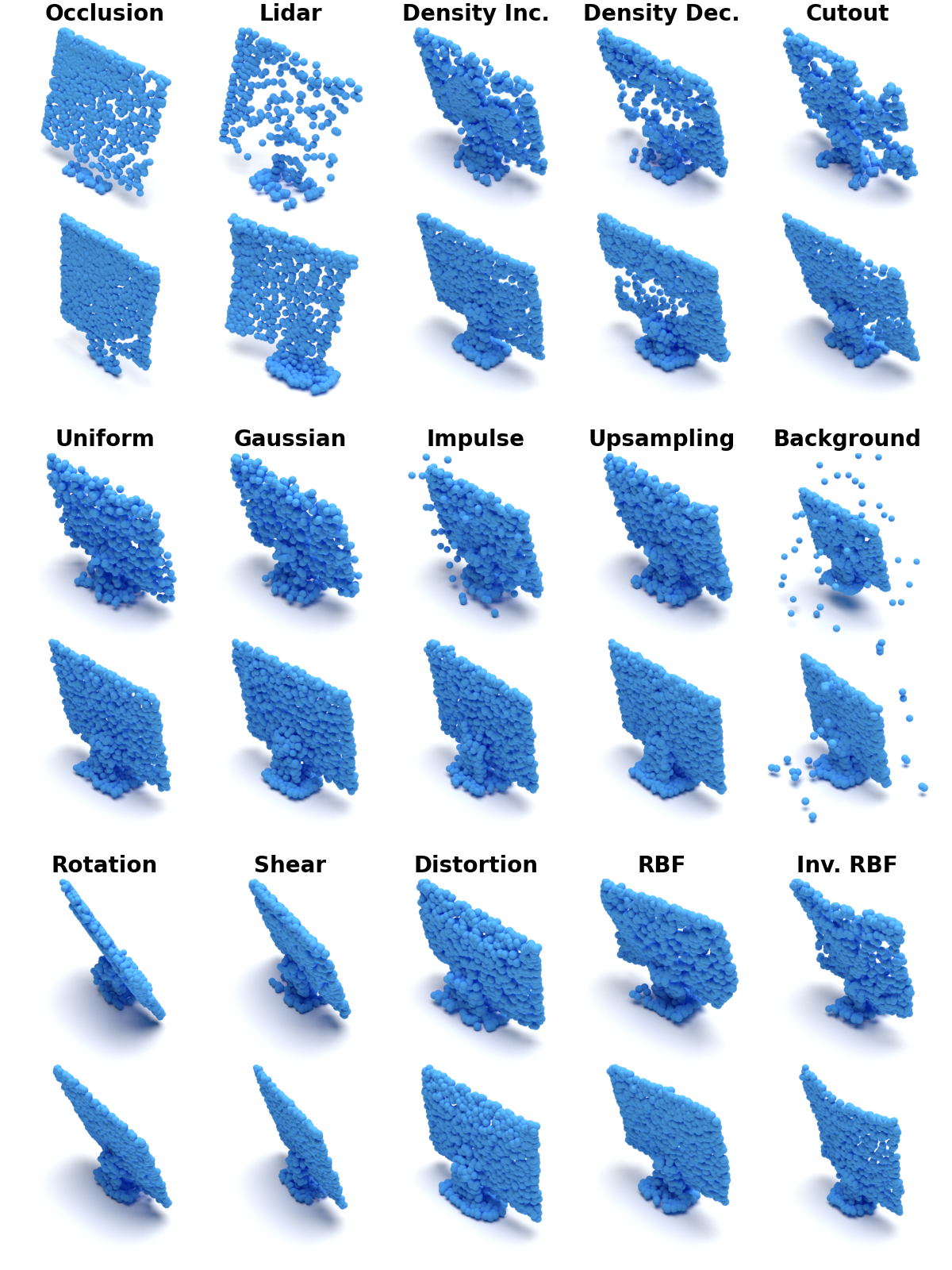}
  \caption{Comparison between the corrupted and adapted point clouds following the application of CloudFixer to 15 different corruption types. The example pertains to a single original point cloud belonging to the monitor class in the test set of ModelNet40-C.}
  \label{fig:examples3}
\end{figure*}

We visualize originally corrupted and translated examples using CloudFixer on ModelNet40-C from \cref{fig:examples1} to \ref{fig:examples3}. % We further depict unadapted and adapted examples across six cross-domain settings in PointDA-10, as shown in \cref{fig:examples6} to \ref{fig:examples11}.
These examples provide additional qualitative analysis, confirming that for various severe corruption types, CloudFixer can truly translate the input point clouds to a clean source domain.

%% file: algorithms/cloudfixer.tex
\begin{algorithm}[ht]
% \footnotesize
    \caption{CloudFixer}\label{alg:cloudfixer}
    \begin{algorithmic}
    \Require Test instance $x\in\mathbb{R}^{N\times3}$, Diffusion model $\epsilon_\theta$, Classifier $f_\psi$, Timestep schedule $(t_{min}, t_{max})$, \# of iteration $S$, Regularization scheduling $\lambda(\cdot)$, Learning rate scheduling $\eta_\texttt{input}(\cdot), \eta_\texttt{model}(\cdot)$, \# of nearest neighbor $k$, \# of vote $K$ 
    \State \LineComment{ Diffusion-Guided Input Adaptation}
    
    %, Timestep interval length $L$ %, Regularization schedule $\lambda_l, \lambda_h$, \# of subpoints $S$
    %\State 
    %\State $N \gets \texttt{len}(c)$
    %\State $c \gets \texttt{FPS}(c, S)$ \Comment{Preprocessing}
    %\State $c, \texttt{mask} \gets \texttt{Duplicate}(c, N-S)$
    %\State $w \gets \texttt{Distance\_Based\_Weights}(x, k)$ %$ , \texttt{mask})$
    \State $\Delta \gets {\bf 0}$, ~$R \gets I $, ~$m \gets 0$
    \While{$m < K$} {\textcolor{lightgray}{\Comment{Conducted as batch processing}}}
    \State $n \gets 0$
    \While{$n < S$}
        % \State $t \gets \texttt{Uniform}[t_{min}, t_{max}]$ %$ + \frac{n}{n_c}(t_l - t_h), t_h + \frac{n}{n_c}(t_l - t_h) + L]$, 
        % \State $\epsilon \gets \mathcal{N}({\bf 0}, I)$
        \State $y[m] \gets R(x + \Delta)$
        \State $t,~\epsilon \gets U[t_{min}, t_{max}],~\mathcal{N}(0, I)$
        \State $y_t \gets \alpha_t y[m] + \sigma_t \epsilon$
        \State $\hat{y} \gets (y_t - \sigma_t \epsilon_\theta(y_t, t)) / \alpha_t$
        \State $w_i \gets 1 / \sum_{j\in kNN(i)} \| x_i - x_j \|_2$, $\forall i\in \{1, \dots, N \}$ 
        \State $w_i \gets w_i / \sum_j w_j$, $\forall i\in \{1, \dots, N \}$ 
        \State $\mathcal{L_\texttt{input}} \gets D(\texttt{stopgrad}(\hat{y}), y[m]) + \lambda(n) \sum_j w_j \|\delta_j\|_2^2)$
        \State $(\Delta, R) \gets (\Delta, R) - \eta_\texttt{input}(n)\nabla_{(\Delta, R)}\mathcal{L}_\texttt{input}$ {\textcolor{lightgray}{\Comment{AdaMax in our implementation}}} %texttt{Optimize}(\Delta, y, \epsilon,  \epsilon_\theta, t)$ \Comment{Eq. \eqref{eq:objective}}
        \State $n \gets n+1$
    \EndWhile\textbf{end while}
    \State $m \gets m+1$
    \EndWhile\textbf{end while}
    
    \State \LineComment{ Online Model Adaptation (CloudFixer-O. Optional)}
    \State~$o \gets 0$,
    \While{$o < O$}
        \State $\mathcal{L}_\texttt{model} \gets \sum_{m=1}^K  KL(f_\psi(x), f_\psi(y[m]))$ % + KL(f_\psi(y[k]), f_\psi(x)) \big)$
        \State $\psi \gets \psi -\eta_\texttt{model}(o)\nabla_{\psi}\mathcal{L}_\texttt{model} $ {\textcolor{lightgray}{\Comment{AdamW in our implementation}}}
        \State $o \gets o+1$
    \EndWhile\textbf{end while}

    % \State $w[\sim\texttt{mask}] \gets 0$
    % \While{$n < n_c+n_f$} \Comment{Fine stage}
    % \State $t \gets \texttt{Uniform}[t_l, t_l + L]$,~ $\epsilon \gets \mathcal{N}({\bf 0}, I)$
    %     \State $y \gets (x + \Delta)R^\top$
    %     \State $\Delta, R \gets \texttt{Optimize}((\Delta, R), y, \epsilon, \epsilon_\theta, t)$ \Comment{Eq. \eqref{eq:objective}}
    %     \State $n \gets n+1$
    % \EndWhile \textbf{end while} 
    % \State $y \gets (x + \Delta)R^\top$ \\
    % \Return $y$
    \end{algorithmic}
\end{algorithm}

%% file: tables/hparam_space.tex
\begin{table*}[t!]
\centering
\caption{Hyperparameter search space of test-time adaptation baseline methods.}
\label{table:tta_hyperparam_searchspace}
% \vspace{-.12in}
\setlength{\columnsep}{5pt}
\renewcommand{\arraystretch}{1.2}
\begin{adjustbox}{width=0.7\linewidth}
\begin{tabular}{l|l}
    \Xhline{3\arrayrulewidth}
    
    \bf Method & \bf Hyperparameter Search Space \\ \Xhline{2\arrayrulewidth}
    
    PL~\cite{pl} & learning rate: $\{10^{-4}, 10^{-3}, 10^{-2}\}$, adaptation steps: $\{ 1, 3, 5, 10\}$ \\ \hline
    
    TENT~\cite{wang2020tent} & learning rate: $\{10^{-4}, 10^{-3}, 10^{-2}\}$, adaptation steps: $\{ 1, 3, 5, 10\}$ \\ \hline

    \multirow{2}{*}{SHOT~\cite{shot}} & learning rate: $\{10^{-4}, 10^{-3}, 10^{-2}\}$, adaptation steps: $\{ 1, 3, 5, 10\}$, \\
    & pseudo-labeling loss weight: $\{0, 0.1, 0.3, 0.5, 1\}$ \\ \hline

    \multirow{2}{*}{SAR~\cite{niu2023towards}} & learning rate: $\{10^{-4}, 10^{-3}, 10^{-2}\}$, adaptation steps: $\{ 1, 3, 5, 10\}$, \\
    & entropy threshold: $\{0.2, 0.4, 0.6, 0.8\}$, epsilon threshold: $\{0.01, 0.05, 0.1\}$ \\ \hline

    DUA~\cite{dua} & adaptation steps: $\{ 1, 3, 5, 10\}$, decay factor: $\{0.9, 0.94, 0.99\}$ \\ \hline
    
    \multirow{2}{*}{LAME~\cite{lame}} & kernel affinity: $\{\text{rbf}, \text{kNN}, \text{linear}\}$, \# of neighbors: $\{ 1, 3, 5, 10 \}$, \\ 
    & max steps: $\{1, 10, 100 \}$ \\ \hline
    
    \multirow{2}{*}{MEMO~\cite{memo}} & learning rate: $\{10^{-6}, 10^{-5}, 10^{-4}, 10^{-3}\}$, adaptation steps: $\{ 1, 2\}$, \\
    & \# of augmentations: $\{16, 32, 64\}$ \\ \hline
    
    DDA~\cite{gao2022source} & guidance weight: $\{3, 6, 9\}$, low pass filtering scale: $\{ 2, 4, 6 \}$ \\\hline
        
    %\multirow{2}{*}{\bf CloudFixer} & learning rate: $\{2 \cdot 10^{-2}, 10^{-1}, 2 \cdot 10^{-1}\}$, lr warmup: $\{0, 0.2, 0.4\}$, \\
    %& $\lambda$: $\{5, 10, 20\}$, $t_{max}$: $\{0.12T, 0.22T, 0.32T\}$ \\
    
    \Xhline{3\arrayrulewidth}
\end{tabular}
\end{adjustbox}
\end{table*}

%% file: tables/pointMLP.tex
\begin{table*}[ht]
    \centering
    \caption{Accuracy on ModelNet40-C using PointMLP under the mild conditions of a batch size of 64 and an i.i.d. test stream except for DDA, MEMO, and CloudFixer which operate with a batch size of 1.}
    \label{table:pointmlp}
    % \vspace{-.15in}
    \setlength{\tabcolsep}{4pt}
    \renewcommand{\arraystretch}{1.3}
    \resizebox{0.99\linewidth}{!}{
        \begin{tabular}{cl|cccccccccccccccc|c}
            \Xhline{3\arrayrulewidth}
            & \multirow{2}{*}{\bf Method} & \multicolumn{5}{c|}{\bf Density Corruptions} &  \multicolumn{5}{c|}{\bf Noise Corruptions} &  \multicolumn{5}{c|}{\bf Transformation Corruptions} & \multirow{2}{*}{\bf Avg.} \\ \cmidrule(lr){3-7} \cmidrule(lr){8-12} \cmidrule(lr){13-17}

            & & \bf OC & \bf LD & \bf DI & \bf DD & \bf CO & \bf UNI & \bf GAU & \bf IP & \bf US & \bf BG & \bf ROT & \bf SH & \bf FFD & \bf RBF & \bf IR & \\ \Xhline{2\arrayrulewidth}

            &Unadapted & 29.01 & 6.11 & 87.92 & 79.70 & 79.13 & 53.12 & 37.96 & 36.53 & 20.82 & 7.24 & 48.37 & 85.37 & 79.21 & 74.55 & 77.87 & 53.53 \\ \cdashline{2-18}

            \multirow{5}{*}{\rotatebox{90}{\textbf{Online}}} &PL~\cite{pl} & 31.04 & 17.54 & 68.35 & 69.85 & 55.63 & 65.56 & 61.02 & 58.51 & 54.62 & 65.48 & 51.05 & 69.89 & 73.26 & 68.72 & 63.33 & 58.26 \\

            &TENT~\cite{wang2020tent} & \textbf{50.36} & 31.69 & \textbf{88.82} & \textbf{85.58} & \textbf{85.25} & 84.89 & 80.11 & 85.01 & 80.63 & \underline{78.08} & \underline{80.79} & \textbf{87.32} & \textbf{85.13} & \underline{85.41} & \underline{85.90} & \underline{78.33} \\

            &SHOT~\cite{shot} & 44.53 & 31.32 & 80.06 & 79.09 & 77.51 & 74.23 & 74.84 & 73.14 & 72.85 & 71.84 & 77.19 & 78.53 & 78.00 & 79.70 & 79.86 & 71.51 \\
            
            &SAR~\cite{niu2023towards} & 44.77 & 28.48 & 31.60 & 43.44 & 30.02 & 30.39 & 38.86 & 43.96 & 76.09 & 18.35 & 49.47 & 29.38 & 46.19 & 52.51 & 42.10 & 40.37 \\
            
            &DUA~\cite{dua} & \underline{47.29} & 30.43 & \underline{88.74} & \underline{84.56} & 83.79 & 82.62 & 78.65 & 81.56 & 78.28 & 67.99 & 76.09 & \underline{85.98} & \underline{83.55} & 83.23 & 84.20 & 75.80 \\ \cline{2-18}
            
            \multirow{3}{*}{\rotatebox{90}{\textbf{Episodic}}} & LAME~\cite{lame} & 28.89 & 5.71 & 88.05 & 80.59 & 79.94 & 52.92 & 36.67 & 34.76 & 19.89 & 8.55 & 48.26 & 85.78 & 79.94 & 75.81 & 77.92 & 53.58 \\

            & MEMO~\cite{memo} & 27.23 & 4.66 & 83.18 & 77.47 & 76.54 & 44.25 & 30.31 & 28.32 & 13.94 & 6.16 & 45.26 & 79.58 & 75.36 & 71.19 & 74.43 & 49.19 \\
            
            & DDA~\cite{gao2022source} & 37.56 & 26.13 & 88.70 & 80.55 & 84.44 & 88.86 & 86.26 & 87.16 & 82.17 & 26.13 & 51.46 & 84.36 & 81.28 & 82.25 & 84.16 & 71.43 \\ \cline{2-18}
            
            & \bf CloudFixer & 34.16 & 32.86 & 84.68 & 75.61 & 78.57 & 87.84 & 87.40 & 88.94 & 86.39 & 67.63 & 75.73 & 81.73 & 78.16 & 78.93 & 81.56 & 74.68 \\
            
            & \bf + Voting ($K=3$) & 34.56 & \underline{33.79} & 85.70 & 77.23 & 80.11 & \underline{89.10} & \underline{87.97} & \underline{89.83} & \underline{87.80} & 69.45 & 80.67 & 82.78 & 79.66 & 80.79 & 82.86 & 76.15 \\
            
            & \bf CloudFixer-O & 44.69 & \textbf{38.65} & 88.25 & 82.78 & \underline{84.97} & \textbf{90.40} & \textbf{89.18} & \textbf{89.91} & \textbf{90.32} & \textbf{82.66} & \textbf{85.33} & \underline{85.98} & 83.47 & \textbf{85.86} & \textbf{87.56} & \textbf{80.67} \\
            \Xhline{3\arrayrulewidth}
        \end{tabular}
    }
\end{table*}

%% file: tables/pointNeXt.tex
\begin{table*}[ht]
    \centering
    \caption{Accuracy on ModelNet40-C using PointNeXt under the mild conditions of a batch size of 64 and an i.i.d. test stream except for DDA, MEMO, and CloudFixer which operate with a batch size of 1.}
    \label{table:pointnext}
    % \vspace{-.15in}
    \setlength{\tabcolsep}{4pt}
    \renewcommand{\arraystretch}{1.3}
    \resizebox{0.99\linewidth}{!}{
        \begin{tabular}{cl|cccccccccccccccc|c}
            \Xhline{3\arrayrulewidth}
            & \multirow{2}{*}{\bf Method} & \multicolumn{5}{c|}{\bf Density Corruptions} &  \multicolumn{5}{c|}{\bf Noise Corruptions} &  \multicolumn{5}{c|}{\bf Transformation Corruptions} & \multirow{2}{*}{\bf Avg.} \\ \cmidrule(lr){3-7} \cmidrule(lr){8-12} \cmidrule(lr){13-17}

            & & \bf OC & \bf LD & \bf DI & \bf DD & \bf CO & \bf UNI & \bf GAU & \bf IP & \bf US & \bf BG & \bf ROT & \bf SH & \bf FFD & \bf RBF & \bf IR & \\ \Xhline{2\arrayrulewidth}
            
            &Unadapted  & 41.41 & 27.95 & 87.84 & 86.18 & 86.06 & 69.12 & 57.90 & 70.58 & 77.02 & 50.77 & 42.50 & 79.01 & 76.45 & 75.04 & 77.55 & 67.03 \\\cdashline{2-18}

            \multirow{5}{*}{\rotatebox{90}{\textbf{Online}}} &PL~\cite{pl} & \underline{51.58} & \underline{45.99} & 88.90 & \textbf{87.36} & 86.87 & 86.51 & 84.81 & 86.91 & 87.40 & 79.13 & 76.78 & \textbf{84.32} & \underline{83.18} & 83.91 & \underline{85.21} & 79.92 \\
            &TENT~\cite{wang2020tent} & 49.68 & 44.85 & \underline{89.22} & 86.91 & 87.20 & 85.25 & 84.08 & 85.45 & 86.83 & 76.50 & 70.66 & 82.74 & 81.73 & 82.41 & 83.51 & 78.47 \\
            &SHOT~\cite{shot} & \textbf{54.13} & \textbf{50.28} & 88.90 & \underline{87.24} & 87.20 & 83.55 & 82.86 & 86.51 & 85.58 & \underline{80.27} & \underline{77.96} & \underline{83.10} & \textbf{84.85} & \underline{84.16} & 84.52 & \underline{80.07} \\
            &SAR~\cite{niu2023towards} & 48.87 & 38.82 & 87.76 & 84.85 & 85.78 & 84.48 & 83.83 & 84.28 & 86.02 & 73.26 & 69.00 & 81.20 & 79.21 & 80.83 & 82.05 & 76.68 \\
            &DUA~\cite{dua} & 49.35 & 44.45 & 88.90 & 87.20 & \underline{87.64} & 84.93 & 83.71 & 86.18 & 86.75 & 76.82 & 70.54 & 82.94 & 81.69 & 82.70 & 83.87 & 78.51 \\\cline{2-18}
            
            \multirow{3}{*}{\rotatebox{90}{\textbf{Episodic}}} &LAME~\cite{lame} & 41.13 & 28.04 & 88.09 & 86.47 & 85.66 & 70.18 & 57.90 & 70.91 & 77.11 & 51.78 & 42.75 & 79.09 & 76.54 & 75.00 & 77.88 & 67.24 \\
            &MEMO~\cite{memo} & 8.83 & 3.12 & 34.40 & 24.88 & 23.10 & 32.13 & 29.46 & 30.51 & 39.10 & 7.13 & 20.79 & 33.10 & 29.21 & 25.85 & 27.07 & 24.58 \\
            & DDA~\cite{gao2022source} & 41.41 & 37.84 & \textbf{90.11} & 85.13 & \textbf{87.76} & \underline{89.55} & \underline{89.06} & \underline{88.98} & \underline{90.32} & 58.43 & 45.06 & 78.00 & 79.29 & 82.01 & 82.86 & 75.05 \\\cline{2-18}
            
            & \bf CloudFixer & 41.65 & 35.01 & 81.93 & 77.96 & 83.23 & 87.60 & 87.80 & 88.49 & 88.94 & 76.42 & 71.47 & 74.96 & 76.46 & 80.96 & 83.43 & 75.75 \\
            & \bf CloudFixer-O & 44.81 & 41.69 & 86.14 & 83.79 & 85.41 & \textbf{89.63} & \textbf{89.99} & \textbf{89.71} & \textbf{91.09} & \textbf{85.13} & \textbf{81.73} & 81.24 & 80.51 & \textbf{84.76} & \textbf{86.63} & \textbf{80.15} \\
            \Xhline{3\arrayrulewidth}
        \end{tabular}
    }
\end{table*}

%% file: tables/pointMAE.tex
\begin{table*}[ht]
    \centering
    \caption{Accuracy on ModelNet40-C using PointMAE under the mild conditions of a batch size of 64 and an i.i.d. test stream except for MATE (bsz. 1), DDA, MEMO, CloudFixer, and CloudFixer-O (bsz. 1) which operate with a batch size of 1.}
    \label{table:pointmae_full}
    % \vspace{-.15in}
    \setlength{\tabcolsep}{4pt}
    \renewcommand{\arraystretch}{1.3}
    \resizebox{0.99\linewidth}{!}{
        \begin{tabular}{cl|cccccccccccccccc|c}
            \Xhline{3\arrayrulewidth}
            & \multirow{2}{*}{\bf Method} & \multicolumn{5}{c|}{\bf Density Corruptions} &  \multicolumn{5}{c|}{\bf Noise Corruptions} &  \multicolumn{5}{c|}{\bf Transformation Corruptions} & \multirow{2}{*}{\bf Avg.} \\ \cmidrule(lr){3-7} \cmidrule(lr){8-12} \cmidrule(lr){13-17}

            & & \bf OC & \bf LD & \bf DI & \bf DD & \bf CO & \bf UNI & \bf GAU & \bf IP & \bf US & \bf BG & \bf ROT & \bf SH & \bf FFD & \bf RBF & \bf IR & \\ \Xhline{2\arrayrulewidth}

            & Unadapted & 34.72 & 15.19 & 85.53 & 78.65 & 74.64 & 61.35 & 51.30 & 55.47 & 51.01 & 16.87 & 31.56 & 67.02 & 61.87 & 58.31 & 62.03 & 53.70 \\\cdashline{2-18} 
            
            \multirow{7}{*}{\rotatebox{90}{\textbf{Online}}} &PL~\cite{pl} & 48.01 & 41.45 & 80.83 & 79.13 & 78.44 & 73.74 & 70.38 & 68.44 & 69.57 & 17.83 & 50.85 & 69.81 & 68.68 & 69.25 & 70.18 & 63.77 \\
            
            &TENT~\cite{wang2020tent} & 47.53 & 41.82 & 82.58 & 79.94 & 79.34 & 74.15 & 73.38 & 69.65 & 69.49 & 17.30 & 51.26 & 71.39 & 69.08 & 70.62 & 71.27 & 64.59 \\
            
            &SHOT~\cite{shot} & \textbf{55.71} & \underline{50.16} & 73.66 & 72.33 & 71.35 & 70.71 & 67.34 & 66.73 & 67.10 & 14.59 & 53.36 & 64.71 & 65.11 & 68.31 & 67.75 & 61.93 \\
            
            &SAR~\cite{niu2023towards} & 47.73 & 45.62 & 82.58 & 80.79 & 79.46 & 75.49 & 72.00 & 69.25 & 69.89 & 10.98 & 51.30 & 70.79 & 70.34 & 70.58 & 72.20 & 64.60 \\
            
            &DUA~\cite{dua} & 50.85 & 45.83 & 84.76 & 83.14 & 82.70 & 77.88 & 75.24 & 73.46 & 73.01 & 19.17 & 55.79 & 74.31 & 72.89 & 73.54 & 75.41 & 67.87 \\
            
            &MATE (bsz. 1)~\cite{mirza2022mate} & 52.55 & 49.85 & 85.66 & 82.54 & 80.63 & 82.86 & 78.97 & 73.62 & 60.29 & 13.05 & 57.54 & 75.89 & 75.53 & 76.05 & 78.44 & 68.23 \\
            
            &MATE (bsz. 64)~\cite{mirza2022mate} & \underline{55.06} & 48.87 & \underline{87.24} & \underline{83.31} & \underline{83.87} & 80.55 & 76.42 & 69.57 & 60.78 & 21.35 & 59.72 & \underline{79.29} & 77.67 & 78.93 & 80.19 & 69.52 \\\cline{2-18}

            \multirow{3}{*}{\rotatebox{90}{\textbf{Episodic}}} &LAME~\cite{lame} & 34.89 & 16.05 & 86.55 & 80.92 & 78.32 & 62.68 & 50.49 & 52.92 & 52.23 & 8.51 & 30.88 & 68.03 & 62.52 & 59.56 & 62.40 & 53.80 \\
            
            &MEMO~\cite{memo} & 33.18 & 14.79 & 85.01 & 78.08 & 76.78 & 59.16 & 47.57 & 49.23 & 48.06 & 6.81 & 29.90 & 67.46 & 60.70 & 56.97 & 59.97 & 51.58 \\
            &DDA~\cite{gao2022source} & 39.91 & 37.07 & 87.07 & 79.70 & 81.16 & 87.60 & 87.20 & 88.09 & 85.13 & 19.89 & 38.37 & 69.98 & 70.83 & 73.70 & 76.30 & 68.13 \\\cline{2-18}
            
            & \bf CloudFixer & 34.24 & 35.62 & 79.62 & 71.15 & 73.54 & 86.71 & 87.16 & 87.72 & 82.25 & 50.36 & 54.05 & 65.40 & 68.60 & 70.66 & 72.45 & 67.97 \\
            
            & \bf CloudFixer-O (bsz. 1) & 47.53 & 48.95 & 85.01 & 81.52 & 83.31 & \underline{88.41} & \underline{88.98} & \underline{88.25} & \underline{88.41} & \underline{59.44} & \underline{75.36} & 75.49 & \underline{78.48} & \underline{84.08} & \underline{84.76} & \underline{77.20} \\
            
            & \bf CloudFixer-O (bsz. 64) & 52.55 & \textbf{53.61} & \textbf{88.17} & \textbf{84.76} & \textbf{86.91} & \textbf{89.91} & \textbf{90.56} & \textbf{90.48} & \textbf{89.79} & \textbf{67.50} & \textbf{78.97} & \textbf{79.66} & \textbf{80.15} & \textbf{85.33} & \textbf{87.07} & \textbf{80.36} \\
            \Xhline{3\arrayrulewidth}
        \end{tabular}
    }
\end{table*}

%% file: tables/ablation_full.tex
\begin{table*}[ht]
    \centering
    \caption{Ablation study conducted on all corruptions of ModelNet40C involves dissecting the core strategies in CloudFixer. This includes parameterization, objective, displacement regularization, forward timesteps, voting, and online adaptation.}
    \label{table:ablation_full}
    % \vspace{-.15in}
    \setlength{\tabcolsep}{4pt}
    \renewcommand{\arraystretch}{1.3}
    \resizebox{0.99\linewidth}{!}{    
        \begin{tabular}{l|cccccccccccccccc|c}
            \Xhline{3\arrayrulewidth}
            \multirow{2}{*}{\bf Setting} & \multicolumn{5}{c|}{\bf Density Corruptions} &  \multicolumn{5}{c|}{\bf Noise Corruptions} &  \multicolumn{5}{c|}{\bf Transformation Corruptions} & \multirow{2}{*}{\bf Avg.} \\ \cmidrule(lr){2-6} \cmidrule(lr){7-11} \cmidrule(lr){12-16}

            & \bf OC & \bf LD & \bf DI & \bf DD & \bf CO & \bf UNI & \bf GAU & \bf IP & \bf US & \bf BG & \bf ROT & \bf SH & \bf FFD & \bf RBF & \bf IR & \\ \Xhline{2\arrayrulewidth}

            Unadapted & 41.09 & 20.02 & 90.03 & \textbf{86.51} & 83.91 & 63.41 & 49.68 & 66.25 & 38.33 & 37.68 & 47.04 & 79.29 & 75.97 & 74.68 & 77.51 & 62.09 \\\cdashline{1-17}

            No Parameterization & 41.49 & 39.02 & 87.96 & 80.75 & 82.13 & 91.21 & 90.92 & 91.73 & \underline{91.81} & 75.64 & 49.79 & 77.63 & 78.16 & 80.87 & 83.42 & 76.17 \\
            Rotation $\rightarrow$ Affine & 40.07 & 37.88 & 86.83 & 79.70 & 82.50 & 90.15 & 89.91 & 90.11 & 87.16 & 74.96 & 66.05 & \textbf{84.32} & 78.32 & 82.66 & 83.59 & 76.95 \\\hline
            
            Squared $\ell_2$ & 40.92 & 37.12 & \underline{90.40} & 84.04 & 83.71 & 84.93 & 82.90 & 88.29 & 71.80 & \underline{76.34} & 78.93 & 81.36 & 76.94 & 80.23 & 82.70 & 76.04 \\
            Diffusion Loss & 39.87 & 33.14 & 84.36 & 76.09 & 73.70 & 88.21 & 87.40 & 82.13 & 86.95 & 39.10 & 59.08 & 75.97 & 71.03 & 70.46 & 72.08 & 69.30 \\\hline
            
            No Reg. & 36.59 & 36.26 & 74.11 & 64.02 & 73.26 & 87.76 & 87.32 & 87.84 & 85.17 & 61.30 & 55.71 & 63.21 & 69.00 & 79.05 & 79.86 & 69.36 \\
            Uniform Reg. & \underline{41.61} & \underline{39.30} & 89.10 & 82.62 & \underline{84.68} & 91.05 & 90.56 & 84.68 & 87.80 & 48.82 & 80.02 & 79.90 & 79.34 & 83.87 & 85.62 & 76.60 \\\hline
            
            $t \sim U[0.01T,0.02T]$ & 41.05 & 21.03 & 89.67 & 83.35 & 82.86 & 86.10 & 83.39 & 81.36 & 84.93 & 54.09 & 58.79 & 80.19 & 45.71 & 40.07 & 40.68 & 64.88 \\
            $t \sim U[0.4T, 0.5T]$ & 26.34 & 17.06 & 45.34 & 42.54 & 48.10 & 55.71 & 54.98 & 62.28 & 38.61 & 58.14 & 60.01 & 51.82 & 36.71 & 37.16 & 39.22 & 44.93 \\\hline
            
            \bf CloudFixer & 41.00 & 38.82 & 87.32 & 80.27 & 83.06 & 91.09 & 90.52 & 90.76 & 89.06 & 75.49 & 81.04 & 78.28 & 78.73 & 82.98 & 85.09 & 78.23 \\
            \bf + Voting ($K=5$) & 41.00 & 38.90 & 88.05 & 80.55 & 84.16 & \underline{91.45} & \underline{91.29} & \textbf{91.98} & 89.79 & \textbf{76.58} & \underline{83.51} & 79.38 & \underline{79.70} & \underline{84.12} & \underline{85.90} & \underline{79.09} \\
            \bf CloudFixer-O & \textbf{46.39} & \textbf{44.94} & \textbf{90.92} & \underline{84.76} & \textbf{86.99} & \textbf{91.94} & \textbf{91.86} & \underline{91.82} & \textbf{92.14} & 74.92 & \textbf{85.98} & \underline{83.83} & \textbf{82.09} & \textbf{86.30} & \textbf{87.40} & \textbf{81.49} \\
            \Xhline{3\arrayrulewidth}
        \end{tabular}
    }
\end{table*}

%% file: tables/metasets.tex
\begin{table*}[ht]
    \centering
    \caption{Accuracy on ModelNet40-C using Point2Vec under the mild conditions of
    a batch size of 64 and an i.i.d. test stream except for MetaSets~\cite{huang2021metasets},
    CloudFixer, and CloudFixer-O.}
    \label{table:point2vec}
    % \vspace{-.15in}
    \setlength{\tabcolsep}{4pt}
    \renewcommand{\arraystretch}{1.3}
    \resizebox{0.99\linewidth}{!}{
        \begin{tabular}{l|cccccccccccccccc|c}
            \Xhline{3\arrayrulewidth}
            \multirow{2}{*}{\bf Method} & \multicolumn{5}{c|}{\bf Density Corruptions} & \multicolumn{5}{c|}{\bf Noise Corruptions} & \multicolumn{5}{c|}{\bf Transformation Corruptions} & \multirow{2}{*}{\bf Avg.} \\ \cmidrule(lr){2-6} \cmidrule(lr){7-11} \cmidrule(lr){12-16}

            & \bf OC & \bf LD & \bf DI & \bf DD & \bf CO & \bf UNI & \bf GAU & \bf IP & \bf US & \bf BG & \bf ROT & \bf SH & \bf FFD & \bf RBF & \bf IR & \\ \Xhline{2\arrayrulewidth}

            Unadapted & 0.4109 & 0.2002 & 0.9003 & 0.8558 & 0.8391 & 0.6341 & 0.4968 & 0.6625 & 0.3833 & 0.3768 & 0.4704 & 0.7929 & 0.7597 & 0.7468 & 0.7751 & 0.6203 \\\cdashline{1-17}

            MetaSets~\cite{huang2021metasets} & \textbf{0.6146} & \underline{0.5057} & \textbf{0.9340} & \textbf{0.9263} & \textbf{0.9246} & 0.8132 & 0.6852 & 0.7459 & 0.5466 & 0.1102 & 0.4413 & \underline{0.8006} & 0.7703 & 0.7318 & 0.7626 & 0.6875 \\

            \textbf{+ CloudFixer} & 0.5239 & 0.4976 & 0.8995 & \underline{0.9024} & 0.9109 & \underline{0.9105} & \underline{0.9080} & \textbf{0.9157} & \underline{0.9055} & \underline{0.6451} & \underline{0.8148} & 0.7950 & \underline{0.7978} & \underline{0.8278} & \underline{0.8359} & \underline{0.8060} \\

            \textbf{+ CloudFixer-O} & \underline{0.5474} & \textbf{0.5255} & \underline{0.9015} & 0.8991 & \underline{0.9125} & \textbf{0.9169} & \textbf{0.9182} & \underline{0.9141} & \textbf{0.9169} & \textbf{0.7098} & \textbf{0.8630} & \textbf{0.8416} & \textbf{0.8432} & \textbf{0.8716} & \textbf{0.8764} & \textbf{0.8305} \\
            \Xhline{3\arrayrulewidth}
        \end{tabular}
    }
\end{table*}

%% file: tables/adv_attack.tex
\begin{table}[htbp]
\centering
\caption{Accuracy of various test-time adaptation methods under adversarial attacks~\cite{pgd} on the original clean test set of ModelNet40 using PointMLP. We further provide the clean (oracle) accuracy without any adversarial attacks.}
\label{tab:adv}
% \vspace{-.1in}
\setlength{\tabcolsep}{10pt}
\renewcommand{\arraystretch}{1.2}
\resizebox{0.3\linewidth}{!}{
% \begin{tabular}{l|cccccccccc|c}
\begin{tabular}{l|c}
    \Xhline{3\arrayrulewidth}
    \textbf{Method} & \textbf{Accuracy} \\ \Xhline{2\arrayrulewidth}
    
    Clean (Oracle) & 93.84 \\
    Adversarial & 11.30 \\ \hdashline
    PL & \underline{42.34} \\
    TENT & 16.29 \\
    SHOT & 35.25 \\
    SAR & 11.59 \\
    DUA & 15.68 \\
    LAME & 11.26 \\
    MEMO & 13.49 \\
    DDA & 39.18 \\ \hline
    \bf CloudFixer & \textbf{79.58} \\

    % \bf Method & Clean (Oracle) & Adversarial & PL~\cite{pl} & TENT~\cite{wang2020tent} & SHOT~\cite{shot} & SAR~\cite{niu2023towards} & DUA~\cite{dua} & LAME~\cite{lame} & MEMO~\cite{memo} & DDA~\cite{gao2022source} & \bf CloudFixer \\ \hline
    % \bf Accuracy & 93.84 & 19.04 & 55.27 & 47.16 & 51.54 & 57.33 & 45.42 & 19.53 & 19.73 & 54.62 & \bf 71.40 \\
    
    \Xhline{3\arrayrulewidth}
\end{tabular}
}
\end{table}

%% file: main.bbl
\begin{thebibliography}{10}
\providecommand{\url}[1]{\texttt{#1}}
\providecommand{\urlprefix}{URL }
\providecommand{\doi}[1]{https://doi.org/#1}

\bibitem{point2vec}
Abou~Zeid, K., Schult, J., Hermans, A., Leibe, B.: Point2vec for self-supervised representation learning on point clouds. In: GCPR (2023)

\bibitem{achituve2021self}
Achituve, I., Maron, H., Chechik, G.: Self-supervised learning for domain adaptation on point clouds. In: WACV (2021)

\bibitem{alexiou2017towards}
Alexiou, E., Upenik, E., Ebrahimi, T.: Towards subjective quality assessment of point cloud imaging in augmented reality. In: IEEE MMSP (2017)

\bibitem{alliegro2021joint}
Alliegro, A., Boscaini, D., Tommasi, T.: Joint supervised and self-supervised learning for 3d real world challenges. In: ICPR (2021)

\bibitem{lame}
Boudiaf, M., Mueller, R., Ben~Ayed, I., Bertinetto, L.: Parameter-free online test-time adaptation. In: CVPR (2022)

\bibitem{cardace2023self}
Cardace, A., Spezialetti, R., Ramirez, P.Z., Salti, S., Di~Stefano, L.: Self-distillation for unsupervised 3d domain adaptation. In: WACV (2023)

\bibitem{shapenet}
Chang, A.X., Funkhouser, T., Guibas, L., Hanrahan, P., Huang, Q., Li, Z., Savarese, S., Savva, M., Song, S., Su, H., et~al.: Shapenet: An information-rich 3d model repository. arXiv preprint arXiv:1512.03012  (2015)

\bibitem{chen20203d}
Chen, S., Liu, B., Feng, C., Vallespi-Gonzalez, C., Wellington, C.: 3d point cloud processing and learning for autonomous driving: Impacting map creation, localization, and perception. IEEE SPM  (2020)

\bibitem{ilvr}
Choi, J., Kim, S., Jeong, Y., Gwon, Y., Yoon, S.: {ILVR:} conditioning method for denoising diffusion probabilistic models. In: ICCV (2021)

\bibitem{scannet}
Dai, A., Chang, A.X., Savva, M., Halber, M., Funkhouser, T., Nie{\ss}ner, M.: Scannet: Richly-annotated 3d reconstructions of indoor scenes. In: CVPR (2017)

\bibitem{fan2022self}
Fan, H., Chang, X., Zhang, W., Cheng, Y., Sun, Y., Kankanhalli, M.: Self-supervised global-local structure modeling for point cloud domain adaptation with reliable voted pseudo labels. In: CVPR (2022)

\bibitem{sam}
Foret, P., Kleiner, A., Mobahi, H., Neyshabur, B.: Sharpness-aware minimization for efficiently improving generalization. In: ICLR (2021)

\bibitem{gao2022source}
Gao, J., Zhang, J., Liu, X., Darrell, T., Shelhamer, E., Wang, D.: Back to the source: Diffusion-driven adaptation to test-time corruption. In: CVPR (2023)

\bibitem{gong2022diffuseq}
Gong, S., Li, M., Feng, J., Wu, Z., Kong, L.: Diffuseq: Sequence to sequence text generation with diffusion models. In: ICLR (2023)

\bibitem{note}
Gong, T., Jeong, J., Kim, T., Kim, Y., Shin, J., Lee, S.J.: Note: Robust continual test-time adaptation against temporal correlation. In: NeurIPS (2022)

\bibitem{guo2020deep}
Guo, Y., Wang, H., Hu, Q., Liu, H., Liu, L., Bennamoun, M.: Deep learning for 3d point clouds: A survey. IEEE TPAMI  (2020)

\bibitem{6drepnet}
Hempel, T., Abdelrahman, A.A., Al-Hamadi, A.: 6d rotation representation for unconstrained head pose estimation. In: ICIP (2022)

\bibitem{ho2022imagen}
Ho, J., Chan, W., Saharia, C., Whang, J., Gao, R., Gritsenko, A., Kingma, D.P., Poole, B., Norouzi, M., Fleet, D.J., et~al.: Imagen video: High definition video generation with diffusion models. arXiv preprint arXiv:2210.02303  (2022)

\bibitem{ddpm}
Ho, J., Jain, A., Abbeel, P.: Denoising diffusion probabilistic models. In: NeurIPS (2020)

\bibitem{ho2022video}
Ho, J., Salimans, T., Gritsenko, A., Chan, W., Norouzi, M., Fleet, D.J.: Video diffusion models. arXiv preprint arXiv:2204.03458  (2022)

\bibitem{e3_diffusion}
Hoogeboom, E., Satorras, V.G., Vignac, C., Welling, M.: Equivariant diffusion for molecule generation in 3d. In: ICML (2022)

\bibitem{huang2021metasets}
Huang, C., Cao, Z., Wang, Y., Wang, J., Long, M.: Metasets: Meta-learning on point sets for generalizable representations. In: CVPR (2021)

\bibitem{huang2022manifold}
Huang, H., Chen, C., Fang, Y.: Manifold adversarial learning for cross-domain 3d shape representation. In: ECCV (2022)

\bibitem{jun2023shape}
Jun, H., Nichol, A.: Shap-e: Generating conditional 3d implicit functions. arXiv preprint arXiv:2305.02463  (2023)

\bibitem{sgem}
Kim, C., Park, J., Shim, H., Yang, E.: {SGEM}: Test-time adaptation for automatic speech recognition via sequential-level generalized entropy minimization. In: INTERSPEECH (2023)

\bibitem{kim2022diffusion}
Kim, G., Shim, H., Kim, H., Choi, Y., Kim, J., Yang, E.: Diffusion video autoencoders: Toward temporally consistent face video editing via disentangled video encoding. In: CVPR (2023)

\bibitem{adamax}
Kingma, D.P., Ba, J.: Adam: A method for stochastic optimization. In: ICLR (2015)

\bibitem{pl}
Lee, D.H., et~al.: Pseudo-label: The simple and efficient semi-supervised learning method for deep neural networks. In: Workshop on challenges in representation learning, ICML (2013)

\bibitem{lehner20223d}
Lehner, A., Gasperini, S., Marcos-Ramiro, A., Schmidt, M., Mahani, M.A.N., Navab, N., Busam, B., Tombari, F.: 3d-vfield: Adversarial augmentation of point clouds for domain generalization in 3d object detection. In: CVPR (2022)

\bibitem{li2022diffusion}
Li, X., Thickstun, J., Gulrajani, I., Liang, P.S., Hashimoto, T.B.: Diffusion-lm improves controllable text generation. In: NeurIPS (2022)

\bibitem{shot}
Liang, J., Hu, D., Feng, J.: Do we really need to access the source data? source hypothesis transfer for unsupervised domain adaptation. In: ICML (2020)

\bibitem{lim2023ttn}
Lim, H., Kim, B., Choo, J., Choi, S.: Ttn: A domain-shift aware batch normalization in test-time adaptation. In: ICLR (2023)

\bibitem{liu2021ttt++}
Liu, Y., Kothari, P., Van~Delft, B., Bellot-Gurlet, B., Mordan, T., Alahi, A.: Ttt++: When does self-supervised test-time training fail or thrive? In: NeurIPS (2021)

\bibitem{liu2023meshdiffusion}
Liu, Z., Feng, Y., Black, M.J., Nowrouzezahrai, D., Paull, L., Liu, W.: Meshdiffusion: Score-based generative 3d mesh modeling. In: ICLR (2023)

\bibitem{Luo_2021_CVPR}
Luo, S., Hu, W.: Diffusion probabilistic models for 3d point cloud generation. In: CVPR (2021)

\bibitem{lyu2021conditional}
Lyu, Z., Kong, Z., Xu, X., Pan, L., Lin, D.: A conditional point diffusion-refinement paradigm for 3d point cloud completion. In: ICLR (2022)

\bibitem{ma2022unsupervised}
Ma, C., Yang, Y., Guo, J., Pan, F., Wang, C., Guo, Y.: Unsupervised point cloud completion and segmentation by generative adversarial autoencoding network. In: NeurIPS (2022)

\bibitem{pointMLP}
Ma, X., Qin, C., You, H., Ran, H., Fu, Y.: Rethinking network design and local geometry in point cloud: A simple residual mlp framework. In: ICLR (2022)

\bibitem{pgd}
Madry, A., Makelov, A., Schmidt, L., Tsipras, D., Vladu, A.: Towards deep learning models resistant to adversarial attacks. In: ICLR (2018)

\bibitem{sdedit}
Meng, C., Song, Y., Song, J., Wu, J., Zhu, J.Y., Ermon, S.: Sdedit: Image synthesis and editing with stochastic differential equations. In: ICLR (2022)

\bibitem{dua}
Mirza, M.J., Micorek, J., Possegger, H., Bischof, H.: The norm must go on: Dynamic unsupervised domain adaptation by normalization. In: CVPR (2022)

\bibitem{mirza2022mate}
Mirza, M.J., Shin, I., Lin, W., Schriebl, A., Sun, K., Choe, J., Possegger, H., Kozinski, M., Kweon, I.S., Yoon, K.J., et~al.: Mate: Masked autoencoders are online 3d test-time learners. In: ICCV (2023)

\bibitem{point-e}
Nichol, A., Jun, H., Dhariwal, P., Mishkin, P., Chen, M.: Point-e: A system for generating 3d point clouds from complex prompts. arXiv preprint arXiv:2212.08751  (2022)

\bibitem{niu2023towards}
Niu, S., Wu, J., Zhang, Y., Wen, Z., Chen, Y., Zhao, P., Tan, M.: Towards stable test-time adaptation in dynamic wild world. In: ICLR (2023)

\bibitem{pointMAE}
Pang, Y., Wang, W., Tay, F.E., Liu, W., Tian, Y., Yuan, L.: Masked autoencoders for point cloud self-supervised learning. In: ECCV (2022)

\bibitem{poole2022dreamfusion}
Poole, B., Jain, A., Barron, J.T., Mildenhall, B.: Dreamfusion: Text-to-3d using 2d diffusion. In: ICLR (2023)

\bibitem{pointnet}
Qi, C.R., Su, H., Mo, K., Guibas, L.J.: Pointnet: Deep learning on point sets for 3d classification and segmentation. In: CVPR (2017)

\bibitem{pointnet++}
Qi, C.R., Yi, L., Su, H., Guibas, L.J.: Pointnet++: Deep hierarchical feature learning on point sets in a metric space. In: NeurIPS (2017)

\bibitem{pointNeXt}
Qian, G., Li, Y., Peng, H., Mai, J., Hammoud, H., Elhoseiny, M., Ghanem, B.: Pointnext: Revisiting pointnet++ with improved training and scaling strategies. In: NeurIPS (2022)

\bibitem{qin2019pointdan}
Qin, C., You, H., Wang, L., Kuo, C.C.J., Fu, Y.: Pointdan: A multi-scale 3d domain adaption network for point cloud representation. In: NeurIPS (2019)

\bibitem{ramesh2022hierarchical}
Ramesh, A., Dhariwal, P., Nichol, A., Chu, C., Chen, M.: Hierarchical text-conditional image generation with clip latents. arXiv preprint arXiv:2204.06125  (2022)

\bibitem{ramesh2021zero}
Ramesh, A., Pavlov, M., Goh, G., Gray, S., Voss, C., Radford, A., Chen, M., Sutskever, I.: Zero-shot text-to-image generation. In: ICML (2021)

\bibitem{saharia2022photorealistic}
Saharia, C., Chan, W., Saxena, S., Li, L., Whang, J., Denton, E.L., Ghasemipour, K., Gontijo~Lopes, R., Karagol~Ayan, B., Salimans, T., et~al.: Photorealistic text-to-image diffusion models with deep language understanding. In: NeurIPS (2022)

\bibitem{shen2022domain}
Shen, Y., Yang, Y., Yan, M., Wang, H., Zheng, Y., Guibas, L.J.: Domain adaptation on point clouds via geometry-aware implicits. In: CVPR (2022)

\bibitem{ddim}
Song, J., Meng, C., Ermon, S.: Denoising diffusion implicit models. In: ICLR (2021)

\bibitem{song2020score}
Song, Y., Sohl-Dickstein, J., Kingma, D.P., Kumar, A., Ermon, S., Poole, B.: Score-based generative modeling through stochastic differential equations. In: ICLR (2021)

\bibitem{modelnet40c}
Sun, J., Zhang, Q., Kailkhura, B., Yu, Z., Xiao, C., Mao, Z.M.: Benchmarking robustness of 3d point cloud recognition against common corruptions. arXiv preprint arXiv:2201.12296  (2022)

\bibitem{sun2020test}
Sun, Y., Wang, X., Liu, Z., Miller, J., Efros, A., Hardt, M.: Test-time training with self-supervision for generalization under distribution shifts. In: ICML (2020)

\bibitem{wang2020tent}
Wang, D., Shelhamer, E., Liu, S., Olshausen, B., Darrell, T.: Tent: Fully test-time adaptation by entropy minimization. In: ICLR (2021)

\bibitem{dgcnn}
Wang, Y., Sun, Y., Liu, Z., Sarma, S.E., Bronstein, M.M., Solomon, J.M.: Dynamic graph cnn for learning on point clouds. In: ACM TOG (2019)

\bibitem{wei2022learning}
Wei, X., Gu, X., Sun, J.: Learning generalizable part-based feature representation for 3d point clouds. In: NeurIPS (2022)

\bibitem{modelnet40}
Wu, Z., Song, S., Khosla, A., Tang, X., Xiao, J.: 3d shapenets for 2.5 d object recognition and next-best-view prediction. arXiv preprint arXiv:1406.5670  (2014)

\bibitem{modelnet}
Wu, Z., Song, S., Khosla, A., Yu, F., Zhang, L., Tang, X., Xiao, J.: 3d shapenets: A deep representation for volumetric shapes. In: CVPR (2015)

\bibitem{xiao2022learning}
Xiao, H., Cheng, M., Shi, L.: Learning cross-domain features for domain generalization on point clouds. In: PRCV (2022)

\bibitem{bn_stats}
You, F., Li, J., Zhao, Z.: Test-time batch statistics calibration for covariate shift. In: ICLR (2022)

\bibitem{zeng2022lion}
Zeng, X., Vahdat, A., Williams, F., Gojcic, Z., Litany, O., Fidler, S., Kreis, K.: Lion: Latent point diffusion models for 3d shape generation. In: NeurIPS (2022)

\bibitem{memo}
Zhang, M., Levine, S., Finn, C.: Memo: Test time robustness via adaptation and augmentation. In: NeurIPS (2022)

\bibitem{egsde}
Zhao, M., Bao, F., Li, C., Zhu, J.: Egsde: Unpaired image-to-image translation via energy-guided stochastic differential equations. In: NeurIPS (2022)

\bibitem{zhou20213d}
Zhou, L., Du, Y., Wu, J.: 3d shape generation and completion through point-voxel diffusion. In: ICCV (2021)

\bibitem{zou2021geometry}
Zou, L., Tang, H., Chen, K., Jia, K.: Geometry-aware self-training for unsupervised domain adaptation on object point clouds. In: ICCV (2021)

\end{thebibliography}
